\documentclass{article} 
\usepackage{iclr2025_conference,times}

\usepackage{amsmath,amsfonts,bm}

\def\eqref#1{equation~\ref{#1}}

\def\1{\bm{1}}

\DeclareMathAlphabet{\mathsfit}{\encodingdefault}{\sfdefault}{m}{sl}
\SetMathAlphabet{\mathsfit}{bold}{\encodingdefault}{\sfdefault}{bx}{n}

\usepackage{hyperref}
\usepackage{url}

\usepackage[utf8]{inputenc} 
\usepackage[T1]{fontenc}    
\usepackage{hyperref}       
\usepackage{url}            
\usepackage{booktabs}       
\usepackage{amsfonts}       
\usepackage{nicefrac}       
\usepackage{microtype}      
\usepackage{caption}
\usepackage{wrapfig}   
\usepackage{subcaption}
\usepackage{amsmath}
\usepackage{enumitem}
\usepackage{xcolor}         

\setlist[enumerate]{nosep, leftmargin=*, label=\textbf{[\arabic*]}, itemsep=2pt}
\usepackage{mdframed}
\newmdenv[
  backgroundcolor=gray!20,
  linecolor=gray!20,
  linewidth=0pt,
  innertopmargin=6pt,
  innerbottommargin=6pt,
  innerleftmargin=6pt,
  innerrightmargin=6pt
]{graybox}
\usepackage{pifont}

\usepackage{amsmath}
\usepackage{amssymb}
\usepackage{mathtools}
\usepackage{amsthm}
\usepackage{dsfont}
\usepackage{multicol}
\usepackage{multirow}
\usepackage{xspace}
\usepackage{subcaption} 
\usepackage[capitalize,noabbrev]{cleveref}

\theoremstyle{plain}

\theoremstyle{definition}

\theoremstyle{remark}

\usepackage[textsize=tiny]{todonotes}

\newcommand{\featuregating}{Feature Gating\xspace}
\newcommand{\grouprouting}{Group Routing\xspace}
\newcommand{\icc}{\texttt{InterpretCC}\xspace}

\title{Intrinsic User-Centric Interpretability through Global Mixture of Experts}

\author{
\hspace{-1mm}Vinitra Swamy \\ 
EPFL \\
\small vinitra.swamy@epfl.ch \\
\And 
\hspace{-2mm}Syrielle Montariol \\
  \hspace{-2mm}EPFL \\
    \hspace{-2mm}\small syrielle.montariol@epfl.ch\\
\And
  \hspace{-5mm}Julian Blackwell \\
    \hspace{-5mm}EPFL \\
    \hspace{-5mm}\small julian.blackwell@epfl.ch\\
\AND
  Jibril Frej \hspace{5mm} \\
  EPFL \hspace{5mm} \\
    \small jibril.frej@epfl.ch\hspace{5mm} \\
  \And
    \hspace{13mm} Martin Jaggi \\
  \hspace{13mm} EPFL \\
      \hspace{13mm} \small martin.jaggi@epfl.ch\\
  \And
    \hspace{15mm} Tanja K\"aser \\
  \hspace{15mm} EPFL \\
    \hspace{15mm} \small tanja.kaeser@epfl.ch\\
}

\iclrfinalcopy 
\begin{document}

\maketitle
\vspace{-5mm}
\begin{abstract}
In human-centric settings like education or healthcare, model accuracy and model explainability are key factors for user adoption. Towards these two goals, intrinsically interpretable deep learning models have gained popularity, focusing on accurate predictions alongside faithful explanations. However, there exists a gap in the human-centeredness of these approaches, which often produce nuanced and complex explanations that are not easily actionable for downstream users. We present \icc (interpretable conditional computation), a family of intrinsically interpretable neural networks at a unique point in the design space that optimizes for ease of human understanding and explanation faithfulness, while maintaining comparable performance to state-of-the-art models. \icc achieves this through adaptive sparse activation of features before prediction, allowing the model to use a different, minimal set of features for each instance. We extend this idea into an interpretable, global mixture-of-experts (MoE) model that allows users to specify topics of interest, discretely separates the feature space for each data point into topical subnetworks, and adaptively and sparsely activates these topical subnetworks for prediction. We apply \icc for text, time series and tabular data across several real-world datasets, demonstrating comparable performance with non-interpretable baselines and outperforming intrinsically interpretable baselines. Through a user study involving 56 teachers, \icc explanations are found to have higher actionability and usefulness over other intrinsically interpretable approaches.
\end{abstract}
\vspace{-3mm}
\section{Introduction}
\vspace{-2mm}
The rise in popularity of neural networks (NNs) over traditionally interpretable models has come with a severe weakness: the lack of transparency of their predictions. Neural networks are considered black-box models due to their high number of parameters and complex operations~\citep{molnar2020interpretable}. Humans cannot understand how neural network decisions are made under the hood; this is a crucial limitation in sensitive applications such as education or healthcare, where models' predictions might impact humans~\citep{conati2018ai, vellido2020importance}.

Towards transparency for neural networks, a range of explainable AI methods have emerged across two main axes: \textit{global} vs. \textit{local} explanations, and \textit{post-hoc} vs. \textit{intrinsic} explanations~\citep{du2019techniques}. \textit{Global} interpretability allows users to understand how the entire model operates by examining its structure and parameters, while \textit{local} interpretability focuses on understanding the rationale behind an individual prediction~\citep{molnar2020interpretable}. Most methods are \textit{post-hoc}, where the explanation is extracted from a model that has already been trained. In contrast, \textit{intrinsic} methods directly incorporate interpretability into the model's structure. 

Popular post-hoc methods include attribution methods like LIME~\citep{lime} or SHAP~\citep{shap},  counterfactual methods like CEM~\citep{counterfactual}, or pattern-based methods like PREMISE~\citep{hedderich2022label}. Post-hoc methods require users to trust the explainer’s approximation of the ground truth explanation (the underlying model's decision process), although they have been shown to be systematically biased and inconsistent~\citep{krishna2022disagreement, swamy-eval}. \textit{In-hoc} explanations are a subset of post-hoc methods that require access to model weights instead of treating the model as a black box~\citep{molnar2020interpretable}. For instance, Grad-CAM~\citep{gradcam} assesses the contribution of a component to the model’s output, and TCAV/DTCAV~\citep{tcav, dtcav} use user-defined concepts to interpret neural network embeddings.

Recent intrinsically explainable model literature has focused on example-based approaches, overwhelmingly for the image modality (\textit{e.g.} B-cos networks~\citep{bohle2022b}, PIP-Net~\citep{nauta2023pip}, ProtoPNet~\citep{chen2019looks}) and \textcolor{black}{less commonly in time-series, tabular, or text modalities}~\citep{sawada2022concept}. Other approaches like NAM~\citep{agarwal2021neural} and EBM~\citep{nori2019interpretml} train a model for each input feature or combination of features and output predictions using scores from these models, requiring a lot of subnetworks when the feature space is large. Research towards interpretable mixture-of-experts models has highlighted a hierarchical neural network structure with subnetworks, combining interpretable experts (\textit{i.e.,} decision trees) with NNs for partially interpretable points~\citep{ismail2023interpretable}, selectively activating experts~\citep{li2022branch}, or extracting automated concepts over the input space for routing~\citep{you2023sum,alvarez2018towards}. All of these approaches, while expressive, are burdened with overly detailed explanations that limit human understandability and actionability.

\begin{figure}[t]
\vspace{-6mm}
    \centering    \includegraphics[width=\linewidth]{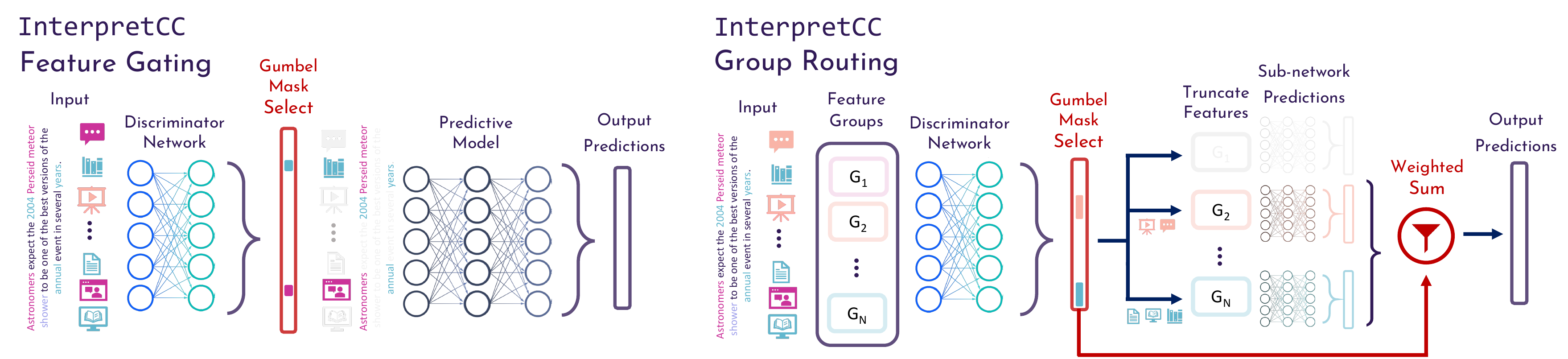}
    \caption{\textbf{\icc Architectures}: \textit{\featuregating (left, individual features)}: (i) All features are input into a discriminator network that outputs a sparse feature activation mask; (ii) Only the features selected via the mask are passed to a predictive network for the final prediction. \textit{\grouprouting (right, pre-defined feature groups)}: (i) Features are statically assigned to distinct groups, with each feature routed to only one group; (ii) Features are input to a discriminator network, generating a sparse group activation mask; (iii) Predictions from activated sub-networks (selected via mask) are aggregated by a weighted sum to produce the final output.}
    \vspace{-7mm}
    \label{fig:arch}  
\end{figure}

In this paper, we therefore present a \textbf{user-centric intrinsically interpretable} framework that achieves faithful local interpretability and provides sparse and actionable human-centric explanations, while maintaining comparable predictive performance to its black-box counterparts. To achieve these goals, we use conditional computation to craft interpretable neural pathways using two different architectures (see Figure \ref{fig:arch}) based on routing through individual features or entire feature groups.

Our models enable statements like the following: \textcolor{black}{``The student’s regularity and video watching behavior were the only two aspects selected as important for the student's prediction of passing the course, and the model did not use any other aspects to make this prediction"}. We refer to interpretability from the users' perspective, focusing on the model's local reasoning for a decision on a specific data point, as opposed to a global understanding of the model's internals. Our models are characterized by sparse explicit routing, truncated feature spaces, and adaptivity per data point. These traits are important for human-centric trustworthiness as they provide clear and concise instance-level explanations~\citep{miller2019explanation, swamy2023future}.

With our family of \icc models, we provide the following contributions:
\begin{enumerate}[itemsep=0em]
    \item \textbf{\icc \featuregating: A simple, interpretable NN architecture} using a gating mechanism to sparsely activate specific features.
    \item \textbf{\icc \grouprouting: An interpretable mixture-of-experts architecture} that uses human-specified group routing to separate the feature space and sparsely activate specific experts.
    \item \textcolor{black}{\textbf{An extension of intrinsic interpretability to multiple human-centric modalities and domains}, focusing on time-series (education), tabular (health, synthetic data), and text (sentiment, news).\footnote{We do not focus on vision datasets since extracting concepts from vision has been well-studied by existing, modality-specific interpretability approaches, \textit{e.g.} \cite{bohle2022b, you2023sum, donnelly2022deformable, thomas2023unlocking}.}}
    \item \textbf{A novel XAI user study} comparing teachers' preferences of interpretable-by-design model explanations towards designing educational interventions for struggling students.
\end{enumerate}

 Across experiments on eight diverse datasets, we show that \icc models perform comparatively to non-interpretable baselines (matching or exceeding 95\% CIs in 15 of 16 comparisons) and outperform intrinsically interpretable baselines (\texttt{ICC} Feature Gating is 9.05\% better on average than SENN Features and 3.27\% better on average than NAM, while \texttt{ICC} Group Routing is on average 5.63\% better than SENN Concepts). Moreover, participants of the user study preferred \texttt{ICC} explanations over baselines in terms of actionability, usefulness, conciseness and trustworthiness.
We provide our code open source: \url{https://github.com/epfl-ml4ed/interpretcc}.

\vspace{-4mm}
\section{Background}
\vspace{-2mm}

\begin{wraptable}{r}{0.5\textwidth}
\vspace{-3mm}
\centering
\resizebox{0.5\textwidth}{!}{%
\begin{tabular}{@{}clllll@{}}
\toprule
\textbf{} & \multicolumn{5}{c}{\textbf{Explanation}} \\ \cmidrule(l){2-6} 
\textbf{Method} & \multicolumn{1}{c}{\textit{Granularity}} & \multicolumn{1}{c}{\textit{Basis}} & \multicolumn{1}{c}{\textit{Faithfulness}} & \multicolumn{1}{c}{\textit{Sparsity}} & \multicolumn{1}{c}{\textcolor{black}{\textit{Stage}}} \\ 
\midrule
\textbf{\begin{tabular}[c]{@{}c@{}}LIME, SHAP\\ (Post-Hoc)\end{tabular}} & Feature & Use all input features & {\color[HTML]{CB0000} Approximation} & {\color[HTML]{009901} Sparse} & {\color[HTML]{CB0000} \begin{tabular}[c]{@{}l@{}}Explanation not \\ used in model\end{tabular}} \\ 
\midrule
\textbf{\begin{tabular}[c]{@{}c@{}}TCAV\\ (In-Hoc)\end{tabular}} & Concept & \begin{tabular}[c]{@{}l@{}}User defines concepts \\ through examples\end{tabular} & {\color[HTML]{F8A102} \begin{tabular}[c]{@{}l@{}}Aligned\\ with concepts\end{tabular}} & {\color[HTML]{009901} Sparse} & {\color[HTML]{F8A102} \begin{tabular}[c]{@{}l@{}}Explain from \\ model internals\end{tabular}} \\ 
\midrule
 & Feature & Use all input features & {\color[HTML]{009901} Guaranteed} & {\color[HTML]{CB0000} Not sparse} & {\color[HTML]{009901} \begin{tabular}[c]{@{}l@{}}Explain \\ then predict\end{tabular}} \\ \cmidrule(l){2-6} 
\multirow{-2}{*}{\textbf{SENN}} & Concept & \begin{tabular}[c]{@{}l@{}}Automated concept \\ selection\end{tabular} & {\color[HTML]{F8A102} \begin{tabular}[c]{@{}l@{}}Aligned \\ with concepts\end{tabular}} & {\color[HTML]{CB0000} Not sparse} & {\color[HTML]{009901} \begin{tabular}[c]{@{}l@{}}Explain \\ then predict\end{tabular}} \\ 
\midrule
\textbf{NAM} & Feature & Use all input features & {\color[HTML]{009901} Guaranteed} & {\color[HTML]{CB0000} Not sparse} & {\color[HTML]{009901} \begin{tabular}[c]{@{}l@{}}Explain \\ then predict\end{tabular}} \\ 
\midrule
 & Feature & Use all input features & {\color[HTML]{009901} Guaranteed} & {\color[HTML]{009901} Sparse} & {\color[HTML]{009901} \begin{tabular}[c]{@{}l@{}}Explain \\ then predict\end{tabular}} \\ \cmidrule(l){2-6} 
\multirow{-3}{*}{\textbf{\begin{tabular}[c]{@{}c@{}}\icc\\(Feature Gating,\\ Group Routing)\end{tabular}}} & Concept & \begin{tabular}[c]{@{}l@{}}User (or LLM) defines \\ groups of features\end{tabular} & {\color[HTML]{009901} Guaranteed} & {\color[HTML]{009901} Sparse} & {\color[HTML]{009901} \begin{tabular}[c]{@{}l@{}}Explain \\ then predict\end{tabular}} \\ 
\bottomrule
\end{tabular}}
\vspace{1mm}
\caption{\textbf{\textcolor{black}{Design} Comparison}: \icc models are at the unique intersection of flexible explanation \textit{granularity} (either features or concepts), guaranteed explanation \textit{faithfulness} to the model's decision process, optimal \textit{sparsity} in the explanation, and \textcolor{black}{explanations used in the prediction (\textit{stage})}. \textit{Basis} describes the foundation of the explanation (\textit{e.g.,} user-defined concepts or raw features). \textcolor{black}{A taxonomy can be found in Appendix \ref{sec:related-work}.} \label{tab:approaches}}
\vspace{-2mm}
\end{wraptable}

\textbf{Architecture Foundations.} \textit{Conditional Computation (CC)} has become widely used to improve the computationally expensive training and inference of large neural networks by activating only parts of the network~\citep{begnio-stochastic, bengio-cc, davis-arel}. 
Inspired by the foundations laid out by CC, mixture-of-expert models have rapidly gained popularity for improving the efficiency of neural networks through activating different expert subnetworks at different layers.
BASE layers~\citep{lewis2021base} direct each token to a designated expert and Switch Transformers~\citep{fedus2022switch} use CC to select one out of 4 feedforward networks across each transformer layer, optimizing computational resources. 
Mixtral~\citep{jiang2024mixtral} is a recent LLM using a mixture of experts to select 2 out of 8 expert networks at each layer, reducing the numbers of active parameters by a factor of 4 compared to training, while allowing each token to have access to all the parameters.  With \icc, we extend a similar routing idea with instance-dependent gating decisions towards an interpretability objective as opposed to only an efficiency or performance objective. 

\textbf{Interpretability Foundations.} Explainability can be integrated into different stages of the modeling pipeline: post-hoc (after model training), in-hoc (requiring model weights), and intrinsic (interpretable by design)~\citep{swamy2023future, molnar2020interpretable}. Our positioning of \textcolor{black}{the design of} \icc in comparison to popular approaches is described in Table \ref{tab:approaches}. Specifically, we categorize the approaches using their basis and the stage they are applied to in the pipeline as well as \textcolor{black}{four key aspects for human-centric explanations: \textit{faithfulness}, the explanation reflects the model behavior with certainty~\citep{lyu2024towards,dasgupta2022framework}; \textit{sparsity}, the model uses a minimal amount of features, optimizing for user understandability and actionability~\citep{sun2024sparse, ayoobi2023sparx}; \textit{predictive stage}, the explanation covers the entirety of what the model uses for prediction~\citep{speith2022review, schwalbe2024comprehensive}; and  \textit{granularity}, the explanation is conveyed in terms of features or concepts~\citep{miller2019explanation, jain2020learning}.}

Post-hoc methods such as LIME~\citep{lime} or SHAP~\citep{shap} approximate what the model finds important, and therefore cannot be considered faithful (cross-feature actions are often not described in the explanation). LIME and SHAP use the full input feature space and can be configured for sparseness, although they have been shown to choose a broad amount of features~\citep{swamy-eval} using default settings. In-hoc interpretability approaches often require users to specify examples to define human-understandable concepts (\textit{e.g.} TCAV~\citep{tcav}, DTCAV~\citep{dtcav}), or use hybrid methods with both human-defined and automated concepts~\citep{sawada2022concept}. In-hoc approaches have limited faithfulness due to the (lack of) completeness of the concepts \textcolor{black}{and do not use explanations directly in the prediction}. Inspired by these approaches, \icc allows users to specify interpretable concepts that are directly useful to them. However, we do not use examples, but instead allow users to specify a grouping over the feature space, achieving both sparsity and \textcolor{black}{explanations used for prediction}.

Initial approaches have explored expert models for intrinsic interpretability. The Interpretable Mixture of Experts (IME) framework~\citep{ismail2023interpretable} uses linear models alongside deep models to provide partially faithful explanations. LIMoE~\citep{mustafa2022multimodal} focuses on visual experts to identify concepts like textures and faces, enhancing interpretability in vision tasks. Similarly, the Sum-of-Parts (SOP) model~\citep{you2023sum} uses sparse feature groups to emphasize the model's reliance on subsets of features for predictions. \textcolor{black}{Approaches in extractive rationale methods and explain-then-predict methods~\citep{jain2020learning, bastings2019interpretable, yu2019rethinking} produce intuitive text explanation guarantees with explanation selection before prediction, but are often not generalizable beyond that modality (see Appendix \ref{sec:rationale}). Few intrinsic approaches use expert knowledge to define concepts directly, instead using prototype examples~\citep{koh2020concept} or rules~\citep{konstantinov2024incorporating}.} The most relevant models to our work are Self-Explaining Neural Networks (SENN)~\citep{alvarez2018towards} and Neural Additive Models (NAM)~\citep{agarwal2021neural}, both neural models similar to \texttt{ICC} as opposed to Explainable Boosted Machines (EBM) which uses trees~\citep{nori2019interpretml}. SENN extracts concepts with prototypical examples and their relevances, but it lacks faithfulness (it cannot explain what is not in a concept), sparsity (it explains all concepts), and concepts do not cover the entire feature space. NAM assigns a model to each feature and combines the outputs linearly, achieving faithfulness but not sparsity, as all features contribute. \texttt{ICC} differs by filtering the feature space instead of using all features (sparsity), using user-defined concepts instead of automated concepts (basis), and assigning each feature to a single group, making feature use explicit \textcolor{black}{(predictive stage)}.

\vspace{-2mm}
\section{Methodology}
\vspace{-2mm}

Given an input $x$, the objective of our approach is to select a sparse subset of $x$ that will be used to predict the output to solve the classification task. We propose two architectures:

\textbf{Feature Gating:} The approach only processes a subset of the features by applying a sparse mask $\mathcal{M}$ on the input $x$ before processing it by a model $f$. The output is given by: $ f(\mathcal{M}(x))$. 

\textbf{Group Routing:} A sparse mixture of models~\citep{fedus2022review} applied on human-interpretable groups of features where each expert is assigned to a group of features: $\sum_{i=1}^{K} G(x)_i \cdot f_i(  \mathcal{M}_{\mathcal{G}}(x)_i)$ where $\mathcal{M}_{\mathcal{G}}(.)_i$ is a sparse mask selecting only the features of group $i$, $f_i$ is the expert model associated with the $i$-th group, and $G(x)_i$ is the output of the gating network for group $i$. If $G(x)_i = 0$, the entire group of features is ignored. 
\vspace{-2mm}
\subsection{Feature Gating}
\vspace{-2mm}

\label{sec:gating}
 \textcolor{black}{\icc \featuregating, shown in Fig.~\ref{fig:arch},} is the first step towards using CC paths for interpretability. The features are first passed through a discriminator network $D$ to select which ones to use for computing the output. 
 The Gumbel Softmax trick~\citep{gumbel} is applied on each dimension of $D(x)$ to select features in a differentiable way (see Appendix \ref{appendix:gumbel} for more details). A feature $j$ is activated (the associated value in the mask is non-zero) if the Gumbel Softmax output exceeds a threshold $\tau$, a hyperparameter. This allows the model to adaptively select the number of features based on each instance, using fewer features for simpler cases and more for complex ones.

The output is computed using a model $f$ on the masked input $\mathcal{M}(x)$. Since the explainability is at the feature level, using a black box model for $f$ does not detract from the interpretability. \textcolor{black}{Notably, \texttt{ICC} FG does not require human specification.}

\vspace{-2mm}
\subsection{Group Routing} 
\vspace{-2mm}
\label{sec:routing} 
We build upon the instance-dependent gating architecture with feature groups. 
As displayed in Figure~\ref{fig:arch}, instead of selecting features individually, the mask is applied to human interpretable groups of features. 
Doing so encourages cross-feature interactions while maintaining a meaningful grouping for human users. 
To select the features belonging to group $i$, we use a binary mask $\mathcal{M}_{\mathcal{G}}(x)_i$ that is computed using human-specified rules. 
In section~\ref{sec:predsetting}, we detail our approach to compute $\mathcal{M}_{\mathcal{G}}(x)_i$ for each dataset used in our experiments.

 \textcolor{black}{\icc \grouprouting} is a sparse mixture of experts utilizing a gating network to assign a weight $G(x)_i$ to each group. 
This process mirrors that of \featuregating, starting with a discriminator network $D_{\mathcal{G}}$ \textcolor{black}{with an input of all features and output of~$K$ dimensions ($K$ is the number of groups).} It then applies the Gumbel Softmax and a threshold $\tau_{\mathcal{G}}$ to each group. The model output is a weighted sum of the output of each expert $f_i$ that only uses the features from the $i$-th group as input. Using our sparsity criteria, we ensure that few groups are used to compute the output, making the \grouprouting intrinsically interpretable at the group level, regardless of the types of models used as experts. \grouprouting enables efficient inference without reducing the number of parameters available during training. During the training phase, we employ soft masking, allowing all weights $G(x)_i$ to remain non-zero, thus granting the model access to every expert. This approach allows the model to leverage the full set of parameters during training, enhancing the training efficiency. At inference time we switch to using a hard mask, making the weights sparse and allowing for interpretability.

\vspace{-4mm}
\section{Experimental Settings}
\vspace{-3mm}
\label{sec:predsetting}

\begin{table}[t]
\vspace{-2mm}
\centering
\resizebox{\textwidth}{!}{
\begin{tabular}{@{}llllllllllll@{}}
\toprule
 & \multicolumn{1}{c}{} & \multicolumn{1}{c}{} & \multicolumn{4}{c}{\textbf{Feature-Based Interpretability}} & \multicolumn{5}{c}{\textbf{Concept-Based Interpretability}} \\ \cmidrule(l){4-7} \cmidrule(l){8-12} 
\multirow{-2}{*}{} & \multicolumn{1}{l}{Dataset} & \multicolumn{1}{c}{\begin{tabular}[c]{@{}c@{}}Non-interpretable \\ Base Module\end{tabular}} & \multicolumn{1}{c}{NAM} & \multicolumn{1}{c}{\begin{tabular}[c]{@{}c@{}}SENN \\ Features\end{tabular}} & \multicolumn{1}{c}{\textcolor{black}{FRESH}} & \multicolumn{1}{c}{\begin{tabular}[c]{@{}c@{}}\icc \\ \featuregating\end{tabular}} & \multicolumn{1}{c}{\begin{tabular}[c]{@{}c@{}}SENN \\ Concepts\end{tabular}} & \multicolumn{1}{c}{\begin{tabular}[c]{@{}c@{}}\icc\\Top K Routing\end{tabular}} & \multicolumn{3}{c}{\begin{tabular}[c]{@{}c@{}}\icc \\ Group Routing\end{tabular}} \\ \midrule
 & DSP & 82.81 \footnotesize{$\pm$ 2.61} & 85.20 \footnotesize{$\pm$ 0.64} & {\color[HTML]{9A0000} 71.70 \footnotesize{$\pm$ 0.95}} &  & {\color[HTML]{009901} 90.75 \footnotesize{$\pm$ 0.01}} & 81.50 \footnotesize{$\pm$ 2.26} & 83.08 \footnotesize{$\pm$ 1.10} & \multicolumn{3}{l}{84.90 \footnotesize{$\pm$ 7.59}} \\
 & Geo & 72.96 \footnotesize{$\pm$ 1.59} & {\color[HTML]{9A0000} 65.12 \footnotesize{$\pm$ 4.07}} & {\color[HTML]{9A0000} 57.90 \footnotesize{$\pm$ 2.69}} & Not & 71.92 \footnotesize{$\pm$ 0.01} & 70.90 \footnotesize{$\pm$ 2.45} & {\color[HTML]{0D0D0D} 80.44 \footnotesize{$\pm$ 3.19}} & \multicolumn{3}{l}{{\color[HTML]{009901} 81.58 \footnotesize{$\pm$ 0.57}}} \\
 & HWTS & 73.93 \footnotesize{$\pm$ 3.76} & 73.11 \footnotesize{$\pm$ 2.13} & 68.63 \footnotesize{$\pm$ 3.78} & Supported & {\color[HTML]{009901} 82.89 \footnotesize{$\pm$ 0.04}} & 75.10 \footnotesize{$\pm$ 11.67} & 72.59 \footnotesize{$\pm$ 2.84} & \multicolumn{3}{l}{78.34 \footnotesize{$\pm$ 0.95}} \\
\multirow{-4}{*}{\textbf{Education}} & VA & 74.90 \footnotesize{$\pm$ 5.28} & 71.39 \footnotesize{$\pm$ 3.38} & 74.37 \footnotesize{$\pm$ 1.11} &  & {\color[HTML]{0D0D0D} 77.80 \footnotesize{$\pm$ 0.01}} & 69.99 \footnotesize{$\pm$ 8.83} & 71.43 \footnotesize{$\pm$ 1.11} & \multicolumn{3}{l}{72.08 \footnotesize{$\pm$ 3.71}} \\ \midrule
\textbf{Health} & \begin{tabular}[c]{@{}l@{}}B. Cancer\end{tabular} & 89.70 \footnotesize{$\pm$ 1.05} & 88.77 \footnotesize{$\pm$ 7.31} & 80.52 \footnotesize{$\pm$ 6.21} & Not Supp.  & {\color[HTML]{9A0000} 78.19 \footnotesize{$\pm$ 3.54}} & {\color[HTML]{9A0000} 85.26 \footnotesize{$\pm$ 1.03}} & {\color[HTML]{333333} 84.66 \footnotesize{$\pm$ 3.02}} & \multicolumn{3}{l}{{\color[HTML]{009901} 94.85 \footnotesize{$\pm$ 1.25}}} \\ \midrule
 & AG News & 89.93 \footnotesize{$\pm$ 3.32} &  &  & 88.73 \footnotesize{$\pm$ 0.69} & 85.72 \footnotesize{$\pm$ 5.31} &  & 87.25 \footnotesize{$\pm$ 2.48} & \multicolumn{3}{l}{90.35 \footnotesize{$\pm$ 1.07}} \\
\multirow{-2}{*}{\textbf{Text}} & SST & 91.12 \footnotesize{$\pm$ 2.03} & \multirow{-2}{*}{\begin{tabular}[c]{@{}l@{}}Not \\ Supported\end{tabular}} & \multirow{-2}{*}{\begin{tabular}[c]{@{}l@{}}Not \\ Supported\end{tabular}} & {\color[HTML]{9A0000} 82.05 \footnotesize{$\pm$ 0.56}} & 88.21 \footnotesize{$\pm$ 3.41} & \multirow{-2}{*}{\begin{tabular}[c]{@{}l@{}}Not \\ Supported\end{tabular}} & 92.98 \footnotesize{$\pm$ 0.88} & \multicolumn{3}{l}{91.75 \footnotesize{$\pm$ 1.86}} \\ \midrule
\textbf{Synthetic} & OpenXAI & 86.67 \footnotesize{$\pm$ 0.31} & 87.85 \footnotesize{$\pm$ 1.31} & 83.67 \footnotesize{$\pm$ 1.86} & Not Supp. & {\color[HTML]{009901} 89.51 \footnotesize{$\pm$ 0.51}} & 84.67 \footnotesize{$\pm$ 4.04} & 90.83 \footnotesize{$\pm$ 1.93} & \multicolumn{3}{l}{89.47 \footnotesize{$\pm$ 2.89}} \\ \bottomrule
\end{tabular}}
\vspace{1mm}
\caption{\label{tab:all_results} 
\textbf{\icc Performance} (avg $\pm$ std) on EDU (balanced accuracy), Text, Health, and Synthetic (accuracy) datasets compared to a non-interpretable baseline and \textcolor{black}{four} intrinsically interpretable baselines. All 95\% CI overlap with the non-interpretable base module for values in black. Colored values indicate significantly higher (green) or lower (red) performance than the base module. The reported \grouprouting results are the best performing \texttt{ICC} variations from Table \ref{tab:group-routing}.}
\vspace{-6mm}
\end{table}
We apply \icc to five domains: education, news classification, sentiment classification, healthcare, and synthetic data covering \textit{Time Series}, \textit{Text}, and \textit{Tabular} inputs; all for classification tasks. 
For \textit{Tabular} features, the input is a vector $x \in \mathds{R}^n$. The mask in the Feature Gating is a sparse vector indicating which tabular feature to use and how important they are (if the weight is non-0) and the groups form a partition over the features. For \textit{Text} features, the input is a sequence of $N$ tokens: $x = [t_1, t_2, \cdots, t_N]$. The mask is a sparse vector that indicates which token to use and each group consists of a subset of the tokens. Finally, we consider \textit{Time Series} of $n$ features across $T$ timesteps: $x \in \mathds{R}^{n\times T}$. We apply the same mask across all time steps for \icc FG and GR.

\textbf{EDU} (\textit{time series, education domain}). We predict student success in the early weeks of four massive open online courses (MOOCs), using students' clickstream data (see Table \ref{tab:EDU_course_stats} in Appendix~\ref{sec:appendix-datasets} for details about the courses). The raw clickstream input is transformed into weekly time-series features that have proven useful for student success prediction in previous literature (\textit{e.g.} total video clicks, forum interactions). We select 45 input features used in multiple studies~\citep{lalle-conati, boroujeni-et-al, chen-cui, marras-vignoud-kaser-2021}. For early prediction, we only use the first $40\%$ of time steps as input.

\textit{Grouping:} To derive human-interpretable concepts from these features, we turn to learning science literature. In \textbf{routing by paper}, we create 10 distinct feature subsets based on handcrafted initial input features from 10 papers, directing each to a specific expert subnetwork. 
For \textbf{routing by pattern}, we organize features according to five learning dimensions identified by~\citep{asadi2023ripple, mejia-patterns}: effort, consistency, regularity, proactivity, control, and assessment-based (see Table~\ref{tab:mejia_patterns} for a detailed feature classification).
Thirdly, \textbf{routing by Large Language Model (LLM)}, uses GPT-4's capabilities to aid humans in feature grouping~\citep{achiam2023gpt}. GPT-4 is prompted as an `expert learning scientist' to group the features into self-regulated behavior categories that are easy to understand, which are then used to separate the features for \icc. More details are included in Appendix~\ref{gpt-4}.

\textbf{AG News and SST} (\textit{text, news and review domains}). For news categorization (\textbf{AG News}), we classify news into four categories (`World', `Sports', `Business', `Sci/Tech') given a title and description of a real-world article~\citep{zhang2015character}. We use 36,000 training samples and 3,000 test samples evenly distributed across categories. For sentiment prediction (Stanford Sentiment Treebank, \textbf{SST}), we use 11,855 sentences from movie reviews labeled by three annotators~\citep{socher2013recursive} and predict a binary sentiment from a sentence fragment. 

\textit{Grouping:} The \texttt{ICC} routing model assigns words to subnetworks using the Dewey Decimal Code (DDC) hierarchy of topics for book classification to create 10 subnetworks (see Table \ref{tab:dewey_decimal_classification}, Appendix \ref{app:fg_text} for more details)~\citep{satija2013theory}. Each word is encoded using SentenceBERT~\citep{reimers2019sentence} and assigned to a subcategory (\textit{i.e.,} the word `school' is assigned to the subcategory `education' under category 300 for `social sciences') and routed to the appropriate parent network.

\textbf{Breast Cancer} (\textit{tabular, healthcare domain}). The Wisconsin Breast Cancer dataset identifies cancerous tissue from fine needle aspirate (FNA) images, with 30 features (10 per cell nucleus) and diagnoses (Malignant: 1, Benign: 0) for 569 patients~\citep{breastcancer}.

\textit{Grouping:} For the grouping logic, we group each cell nucleus in a separate subnetwork, enabling features representing the same part of the tissue sample to be considered together.

\textbf{Synthetic Dataset} (\textit{tabular}) We use OpenXAI's synthetic dataset~\citep{agarwal2022openxai}, which includes ground truth labels and explanations, indicating the subset of features influencing each label. This dataset comprises of 5000 samples, 20 continuous features, and two classes. It was created using the \textit{SynthGauss} mechanism from five cluster neighborhoods (1000 points for each cluster), ensuring three desirable properties for assessing explanations: (1) feature independence, (2) unambiguous, well-separated local neighborhoods, and (3) an explanation for each instance.

\textit{Grouping:} We group the feature space by assigning each feature to a cluster neighborhood based solely on the distribution of the training data. The average absolute value of each feature for each cluster is calculated, and the highest feature-cluster value determines the assignment.

\vspace{-3mm}
\section{Experimental Results}
\label{sec:results}
\vspace{-3mm}

Through the following three experiments, we demonstrate that our \icc models do not compromise \textit{performance} compared to black-box models and provide explanations that are \textit{faithful} as well as \textit{human-centered}. \textcolor{black}{\icc is designed for data that has meaning for humans (\textit{i.e.,} interpretable features or meaningful raw data like text or lab measurements); however, we demonstrate it is also performant on a synthetic dataset with no interpretable features.}

\textbf{Experimental Setup.} We run hyperparameter tuning and three different random seeds for each reported model (reproducibility details in Appendix \ref{sec:reproduce}). Since \textbf{EDU} MOOC courses have a low passing rate (below 30\%), and thus the dataset has a heavy class imbalance, we use balanced accuracy for evaluation. The other datasets are more balanced (\textbf{AG News}, \textbf{SST}, \textbf{Breast Cancer}, \textbf{Synthetic}), hence we use accuracy as our evaluation metric. We perform an 80-10-10 train-validation-test data split stratified on the output label, to conserve the class imbalance in each subset. In addition to \icc \featuregating and \grouprouting, we also employ a \icc Top-K expert network solution with k=2 for group routing. This approach is similar to existing mixture-of-expert approaches~\citep{jiang2024mixtral, li2022branch}, except that their models make a layer-wise expert choice, which significantly reduces interpretability, while we make a global expert choice.

\textbf{Base Prediction Module.} We choose simple yet performant predictive modules reported in previous literature to isolate the difference in performance due to the interpretable architectures. For the EDU data, previous works use BiLSTMs on student behavioral data for best predictive performance~\citep{swamy-eval, marras-vignoud-kaser-2021}. Thus, for comparative benchmarking, the most performant BiLSTM setting is used as a baseline model~\citep{swamy2022meta}. For the AG News and SST datasets, we use fine-tuned \texttt{DistilBERT}\footnote{\url{https://huggingface.co/distilbert/distilbert-base-uncased}} variations as baselines, also used in related works~\citep{yang2019xlnet, hf_canonical_model_maintainers_2022}. For the Breast Cancer dataset, we use a fully connected network as reported by~\cite{agarap2018breast}. 

\textbf{Interpretable Baselines. }
In addition to non-interpretable baselines, we compare \texttt{ICC} to \textcolor{black}{three intrinsically interpretable methods: SENN~\citep{alvarez2018towards}, NAM~\citep{agarwal2021neural}, and FRESH~\citep{jain2020learning}.}
SENN generates concept-based explanations and is designed to learn \textit{"interpretable basis concepts"} in parallel with the model optimization. Explanations consist of the concepts most similar to the input. 
NAM, a General Additive Model~\citep{hastie2017generalized}, uses individual neural networks for each input feature to calculate feature weights. The model's output is the sum of these weights, and explanations are given by displaying the feature weights. \textcolor{black}{The SENN Features architecture has three models working together, NAM has one subnetwork for each feature, and \texttt{ICC FG} has one model with two parts trained end-to-end. At inference, SENN and NAM also assign a score to each feature, which are then aggregated; \texttt{ICC} leverages cross-feature interactions with one score per model. FRESH is an extractive rationale architecture with three models: \textit{supp} generates importance scores, \textit{ext} selects a contiguous text mask, and \textit{pred} applies this mask on the inputs to make predictions. Designed for text data, FRESH's contiguous explanations are unsuitable for modalities like tabular or time-series data, where feature proximity lacks relevance.}
\vspace{-2mm}

\subsection{Exp. 1: \icc does not compromise on performance}
\label{sec:performance}
Table \ref{tab:all_results} shows the models' metrics (average accuracy, standard deviations, and $95\%$ confidence intervals) across three iterations of model training for all eight datasets for \icc Feature Gating, Group Routing, and Top-K Routing, as well as baselines of NAM, SENN Features, SENN Concepts, and a non-interpretable model (reflecting post-hoc explainer performance). We do not report results for text features with NAM and SENN, since the architecture change required to do so would no longer respect the original model design\footnote{For NAM, one network (LLM) would be required per word, as the words are distinct for each instance; it does not support text grouping. For SENN, LMs would need to embed each word, then simultaneously be trained to represent concepts and pick representative words, with new metrics for choosing prototypical examples.}; \textcolor{black}{instead we report text results for FRESH}. \textcolor{black}{Additional sensitivity and architecture experiments can be found in Appendix \ref{appendix:sparsity}. These experiments show that while the performance of \icc has overlapping 95\% CIs while changing parameters, certain parameter settings have higher variability than others. For both education and health tasks, a $\tau$ of 10 and a Gumbel-Softmax threshold of around 0.7 to 0.8 are performant, sparse in activated features, and relatively stable.}

\icc \featuregating statistically significantly improves performance with respect to the non-interpretable base prediction module for two EDU courses (DSP, HWTS) and the synthetic dataset. It shows comparable performance on all other datasets (indicated by the overlapping CIs) except the Breast Cancer dataset, suggesting that a higher number of the available features is necessary for performant prediction. The intrinsically interpretable baselines \textcolor{black}{SENN, NAM, and FRESH never outperform the base model, and always under-perform or perform comparably to \icc}.

\begin{figure}[t]
\vspace{-4mm}
    \centering
    \begin{minipage}[b]{0.55\textwidth}
    \centering
    \resizebox{\textwidth}{!}{
\begin{tabular}{@{}lllll@{}}
\toprule
\multicolumn{1}{c}{\multirow{2}{*}{\textbf{Dataset}}} & \multicolumn{1}{c}{\multirow{2}{*}{\textbf{Baseline}}} & \multicolumn{3}{c}{\textbf{\icc Group Routing}} \\ \cmidrule(l){3-5} 
\multicolumn{1}{c}{} & \multicolumn{1}{c}{} & \multicolumn{1}{c}{\textit{Paper}} & \multicolumn{1}{c}{\textit{Pattern}} & \multicolumn{1}{c}{\textit{GPT-4}}\\ \midrule 
\textbf{DSP} & 82.81 \small{$\pm$ 2.61} & 82.37 \small{$\pm$ 6.27} & 82.29 \small{$\pm$ 3.72} & \textbf{84.90 \small{$\pm$ 7.59}}  \\ 
\textbf{Geo} & 72.96 \small{$\pm$ 1.59} & 69.64 \small{$\pm$ 1.23} & \textbf{81.58 \small{$\pm$ 0.57}} & 81.19 \small{$\pm$ 1.53} \\ 
\textbf{HWTS} & 73.93 \small{$\pm$ 3.76} & \textbf{78.34 \small{$\pm$ 0.95}} & 72.34 \small{$\pm$ 2.77} & 75.12 \small{$\pm$ 4.17}  \\ 
\textbf{VA} & 74.90 \small{$\pm$ 5.28} & 69.88 \small{$\pm$ 2.93} & \textbf{72.08 \small{$\pm$ 3.71}} & 70.98 \small{$\pm$ 2.77} \\ \midrule 
\textbf{Average} & 76.65 \small{$\pm$3.31} & 75.56 \small{$\pm$2.85} & 77.57 \small{$\pm$2.69} & \textbf{78.05 \small{$\pm$4.01} } \\ 
\bottomrule
\end{tabular}
}
\captionof{table}{\textbf{\icc Group Routing Performance}: balanced accuracy (average $\pm$ std) on routing strategies (paper, pattern, GPT-4) for the EDU datasets in comparison to the non-interpretable baseline.}
\label{tab:group-routing}
    \end{minipage}
    \hfill
    \begin{minipage}[b]{0.42\textwidth}
        \centering
        \includegraphics[width=\linewidth]{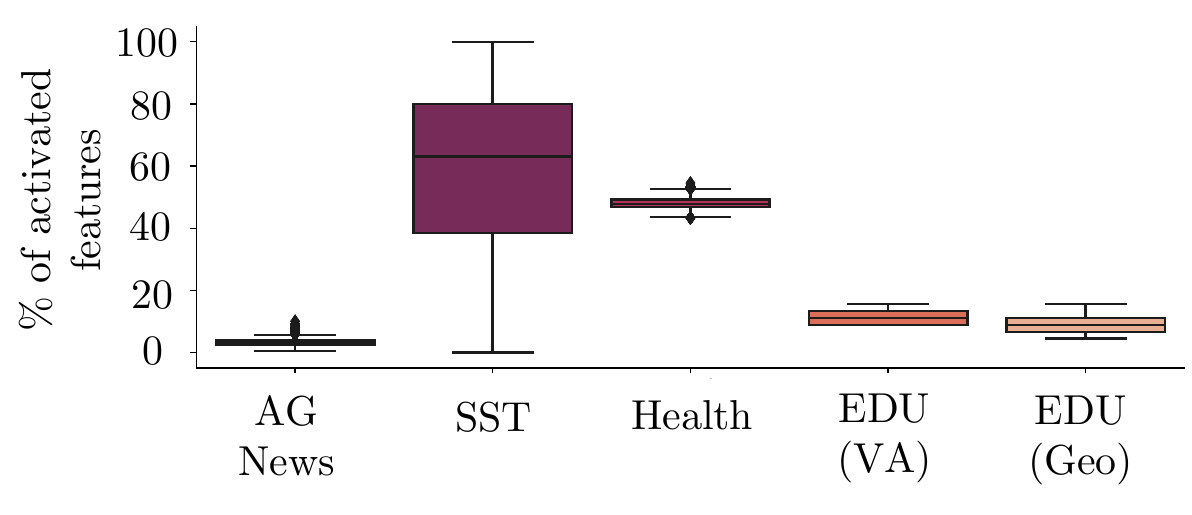}
    \caption{\textbf{\icc Feature Gating Sparsity:} \% of features activated per data point across five representative datasets.}
    \label{fig:sparsity}
    \end{minipage}
\vspace{-4mm}
\end{figure}

\icc \grouprouting outperforms (for the Geo course and Breast Cancer dataset) or performs comparably ($95\%$ CI overlap) to the fine-tuned, non-interpretable base module. It also consistently performs at least comparatively to SENN which, besides never outperforming the base model, has a relatively high variance. 
We further observe that the selected grouping method impacts performance (Table \ref{tab:group-routing}). We achieve a $10\%$ increase in performance compared to the base model when grouping using patterns or GPT-4 for the Geo course. On average, over the four EDU courses, the automated LLM grouping and the pattern-based human-defined grouping perform comparably, showing that using automated grouping methods does not mean compromising on performance.

\vspace{1mm}
\begin{graybox} \icc performs comparably to black-box models and outperforms intrinsically interpretable baselines across diverse benchmarks. \end{graybox}

\subsection{Exp. 2: \icc provides faithful and user-friendly explanations}
\label{sec:faithfulness-sparsity}
\begin{figure}[t]
\vspace{-5mm}
    \centering
    \begin{minipage}[b]{0.34\textwidth}
        \centering \includegraphics[width=\linewidth, trim={5 5 5 5}, clip]{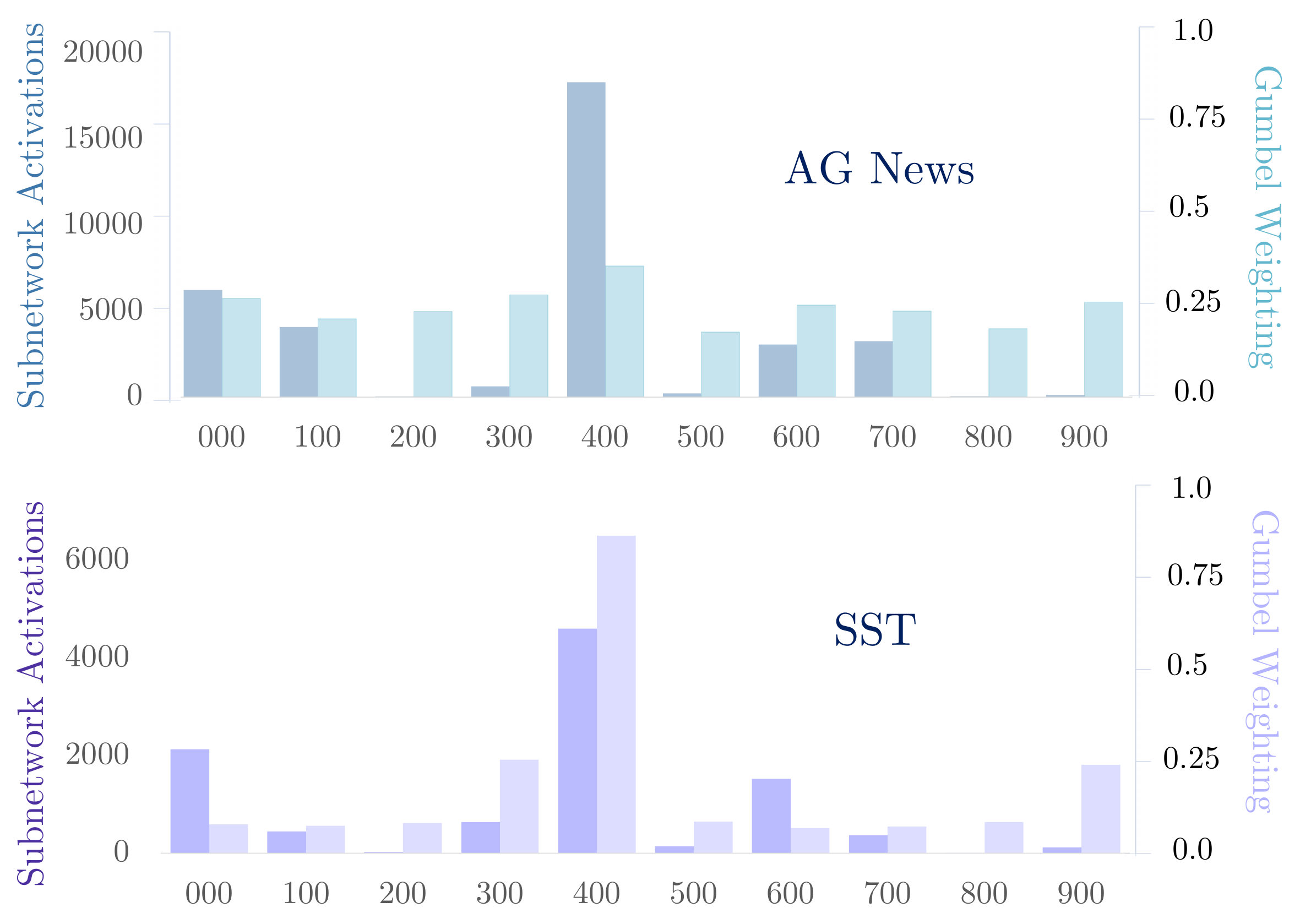}
        \caption{\textbf{AG News} and \textbf{SST}: \# of ICC subnetwork activations (left) vs. avg. activation weights (right), grouped by subnetworks based on the Dewey Decimal Code.}
        \label{fig:ag-news-comparison}
    \end{minipage}
    \hfill
    \begin{minipage}[b]{0.63\textwidth}
\centering
    \resizebox{\textwidth}{!}{
\begin{tabular}{@{}llllc@{}}
\toprule
\multicolumn{1}{c}{\multirow{2}{*}{\textbf{Model}}} &
  \multicolumn{3}{c}{\textbf{Relationship to Underlying Data Patterns}} &
  \textbf{Relationship to Model} \\ \cmidrule(l){2-5} 
\multicolumn{1}{c}{} &
  \multicolumn{1}{c}{\textit{\begin{tabular}[c]{@{}c@{}}Ground Truth \\ Alignment (GTA) \end{tabular}}} &
  \multicolumn{1}{c}{\textit{\begin{tabular}[c]{@{}c@{}}Rank \\ Agreement (RA) \end{tabular}}} &
  \multicolumn{1}{c}{\textit{\begin{tabular}[c]{@{}c@{}}Feature \\ Agreement (FA) \end{tabular}}} &
  \textit{\begin{tabular}[c]{@{}c@{}}Ground Truth \\ Faithfulness (GTF) \end{tabular}} \\ \midrule
\texttt{ICC} FG        & \textbf{94.84 ± 4.11} & \textbf{60.00 ± 12.65} & 87.99 ± 16.00 & always \textbf{100} \\
\texttt{ICC} GR        & 89.51 ± 0.56 & 44.00 ± 14.97 & 76.00 ± 23.32 & always \textbf{100} \\
SENN Features & 85.83 ± 2.56 & 27.99 ± 9.80  & 88.01 ± 9.79  & always \textbf{100} \\
SENN Concepts & 65.19 ± 9.23 & 52.00 ± 20.39 & 80.00 ± 21.90 & always \textbf{100} \\
NAM           & 87.39 ± 2.45 & 40.00 ± 17.89 & 76.00 ± 14.97 & always \textbf{100} \\
IG            & 87.28 ± 1.72 & 56.00 ± 14.97 & \textbf{91.99 ± 16.00} & $\leq 100$          \\
LIME          & 84.75 ± 1.83 & 44.00 ± 23.32 & 64.00 ± 14.97 & $\leq 100$          \\
SHAP          & 83.47 ± 1.03 & 36.00 ± 8.00  & 52.00 ± 20.39 & $\leq 100$          \\ \bottomrule
\end{tabular}}
    \captionof{table}{\textbf{Synthetic faithfulness evaluation} across eight models on explanations using metrics presented in OpenXAI~\citep{agarwal2022openxai}. GTA, RA, and FA (K=5) refer to the explanations' relationship to underlying patterns in the synthetic data. \textcolor{black}{GTF refers to bounds of the explanations' relationship to the model's decision process.} All 95\% CIs overlap between the best performing intrinsically interpretable and non-interpretable model.}
    \label{tab:synth}    
\end{minipage}
    \vspace{-6mm}
\end{figure}

Table \ref{tab:synth} showcases the faithfulness of \icc models in comparison to intrinsically interpretable models SENN and NAM, as well as three post-hoc explainers on top of a non-interpretable model (Integrated Gradients (IG)~\citep{sundararajan2017axiomatic}, LIME~\citep{lime}, and SHAP~\citep{shap}). We examine the relationship of the explanations to underlying data patterns from the synthetic dataset as well as to the ground truth of the underlying model (see Appendix \ref{app:additional-faithfulness} for detailed descriptions of the metrics). All 95\% CI overlap in Ground Truth Alignment (how close the explanation is to the underlying synthetic data feature patterns, similarly to~\citep{liu2021synthetic}), Rank Agreement (RA) and Feature Agreement (FA) from ~\cite{agarwal2022openxai}, except for SENN Concepts performing statistically significantly worse than \icc Feature Gating in GTA.  This indicates that \texttt{ICC}'s identification of important features, and of their order of importance, is close to the underlying data patterns and is either on par or better than both interpretable and non-interpretable models. In terms of predictive performance on the synthetic dataset (last row of Table \ref{tab:all_results}), \texttt{ICC} holds the top three best performing models (89.47, 89.51, 90.83) and the \texttt{ICC} FG variation is significantly more performant than the non-interpretable base model, demonstrating the models' ability to capture signal on this dataset.

Our models provide concise and hence user-friendly explanations through sparse feature (group) activations~\citep{miller2019explanation}. Figure \ref{fig:sparsity} shows the percentage of activated features for \texttt{ICC} \featuregating. For \textbf{EDU}, only about 10\% of the 45 features are activated with low variance. For \textbf{Breast Cancer}, 39.7\% of the features are activated per data point. Unlike other datasets, text-based datasets have a variable number of features (words). In \textbf{AG News}, with an average of 35 words per article, only a small percentage is activated, while in \textbf{SST}, which has shorter sentences (7 words on average), 59.8\% of features are selected with high variance. This achieved sparsity, especially in \textbf{EDU} and \textbf{AG News}, highlights the most important features. This contrasts with post-hoc explainers, which tend to select a broader range of features (\textit{e.g.,} LIME and SHAP for EDU~\citep{swamy-eval}).

 \texttt{ICC} \grouprouting activates different subnetworks with different weights for each data point. Figure \ref{fig:ag-news-comparison} illustrates the number of activations and the average weight for each subnetwork for the text data sets (see Appendix \ref{app:group_routing} for detailed analysis on the Breast Cancer and EDU datasets). For \textbf{AG News} (Figure \ref{fig:ag-news-comparison} top), the average activation weight is similar across all subnetworks (min $0.10$, max $0.21$). However, some subnetworks are activated much more frequently (400 - Language: $18,335$ times). This indicates that most data points will be routed through the same subset of subnetworks, while the remaining subnetworks are important for specific data points only. \textbf{SST} (Figure \ref{fig:ag-news-comparison} bottom) shows similar subnetwork activation patterns. However, in contrast to \textbf{AG News}, the distribution of average weights is not uniform: only three networks are activated with weights larger than $0.15$. We suspect the high weights for subnetwork 400 in Figure \ref{fig:ag-news-comparison} reflect words that DDC has little relation to in the SentenceBERT embedding space.

Examples of EDU domain explanations are in Appendix \ref{app:user-study}, Figures \ref{fig:userstudy-fg} and \ref{fig:userstudy-pr}, with additional AG News examples in Figure \ref{fig:text-example}. We also compare \texttt{ICC} behavior across three grouping strategies for the DSP course (Appendix \ref{appendix:edu_dsp_gr}) and analyze network sparsity at different prediction horizons (\ref{appendix:edu_dsp_sparsity}). Lastly, we show the variation in feature group selection across MOOCs, highlighting \texttt{ICC}'s adaptability regardless of grouping method (Appendix \ref{app:fg_allcourses}).

\vspace{1mm}
\begin{graybox} \icc provides sparse and hence user-friendly explanations, while not compromising on explanation faithfulness. \end{graybox}

\subsection{Exp. 3: \icc explanations are preferred by humans}
\label{sec:user-study}

\textbf{Setting.} To validate the user-centeredness of our approach, we conducted a user study comparing \icc's explanations with the ones from other intrinsically interpretable methods, SENN and NAM. We focused on the education domain and time series input type, using the DSP course in the EDU dataset. We trained \icc \featuregating and \grouprouting (with pattern-based feature grouping strategy, since it is heavily grounded on expert knowledge) as well as SENN and NAM and randomly selected four test samples (\textit{i.e.,} four students) for prediction. In designing the study, we conducted four sequential pilots with eight learning scientists who were unaware of which method we presented in this paper. All explanations were simplified for a non-technical audience and followed consistent templates.

We recruited 56 teachers using \texttt{Prolific}, (see Appendix \ref{app:user-study} for detailed information about the participants' demographics and backgrounds (Figure \ref{fig:userstudy_demo}) and the content of the study). We showed them each model's prediction of the student's success or failure along with its explanation. The explanations were given as a short text and a graph showing the features and concepts used by the model. Note that the choices we made for the presentation of the explanations might have an influence on the participants' perception of the explanations. Examples of how the explanations were presented and a discussion on this limitation can be found in Appendix \ref{app:user-study}, Figures \ref{fig:userstudy-fg}, \ref{fig:userstudy-pr}, \ref{fig:userstudy-senn} and \ref{fig:userstudy-nam}. With our study design, we aimed to highlight each framework's strengths without excessive post-processing. For instance, keeping only the top five features of NAM would be unfair, as other features also contribute to its predictions. Instead, we emphasized the top five features that each contribute positively, negatively, or not at all to the prediction, highlighting its advantage (distinct insights into feature impact) over \icc, while ensuring sparsity. We significantly post-processed the explanations of SENN and NAM to provide them in a format understandable for a non-technical audience and iterated on the visualizations using a human-centered design process~\citep{cooley2000human}. We asked participants to compare these explanations according to five criteria (aligned with~\citep{frej2024finding}), and to rank the criteria in terms of importance: \textbf{Usefulness}: This explanation is useful to understand the prediction. \textbf{Trustworthiness}: This explanation lets me judge if I should trust the model. \textbf{Actionability}: This explanation helps me know how to give feedback to the student.\textbf{Completeness}: This explanation has sufficient detail to understand why the prediction was made. \textbf{Conciseness}: Every detail of this explanation is necessary. 

\begin{figure}[h]
\vspace{-3mm}
    \centering
    \begin{minipage}[t]{0.45\textwidth}
        \centering
        \vspace{-20mm}
        \resizebox{\linewidth}{!}{%
        \setlength{\tabcolsep}{3pt}
        \begin{tabular}{@{}l*{5}{r}@{}} 
        \toprule
        {} & \textbf{NAM} & \textbf{SENN}  & \textbf{\texttt{ICC GR}} & \textbf{\texttt{ICC FG}} & \textbf{Weight} \\
        \midrule
        Usefulness       &  3.25 \footnotesize{$\pm$0.98} &   3.3~~ \footnotesize{$\pm$1.11} &  3.53 \footnotesize{$\pm$1.11} &  \textbf{3.88 \footnotesize{$\pm$0.94}} &     0.28 \\
        Trustworthiness  &  3.28 \footnotesize{$\pm$0.93} &  3.64 \footnotesize{$\pm$0.92} &  3.36 \footnotesize{$\pm$1.06} &  \textbf{3.78 \footnotesize{$\pm$0.9~~}} &     0.23 \\
        Actionability    &  3.08 \footnotesize{$\pm$0.96} &  3.25 \footnotesize{$\pm$1.06} &  3.37 \footnotesize{$\pm$1.04} & \textbf{3.77 \footnotesize{$\pm$0.95}} &     0.21 \\
        Completeness     &  3.18 \footnotesize{$\pm$1.02} & \textbf{3.76 \footnotesize{$\pm$1.09}} &   3.1~~~\footnotesize{$\pm$1.19} &  3.67 \footnotesize{$\pm$1.07} &     0.16 \\
        Conciseness      &  3.13 \footnotesize{$\pm$1.06} &  2.82 \footnotesize{$\pm$1.31} & \textbf{3.72 \footnotesize{$\pm$1.06}} &  3.68 \footnotesize{$\pm$1.05} &     0.12 \\
        \midrule
        Global &   3.2~~~\footnotesize{$\pm$0.81} &  3.38 \footnotesize{$\pm$0.85} &  3.41 \footnotesize{$\pm$0.88} &  \textbf{3.78 \footnotesize{$\pm$0.77}} &      \\
        \bottomrule
        \end{tabular}}
        \caption{\textbf{Model score for each \textcolor{black}{user study} criterion} (average $\pm$ std) and criteria weight according to \textcolor{black}{users'} ranking. All scores range from 1 (lowest) to 5 (highest).}
        \label{tab:userstudy_res}
    \end{minipage}
    \hfill
    \begin{minipage}[t]{0.52\textwidth}
        \centering
        \resizebox{\linewidth}{!}{%
        \includegraphics[trim={3mm 3mm 1mm 3mm},clip]{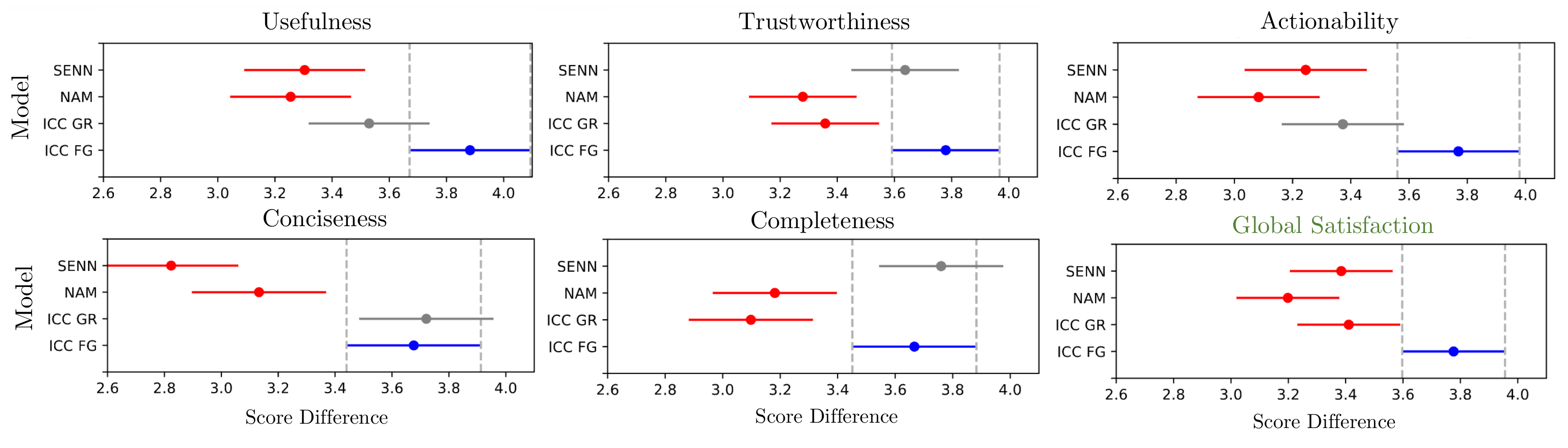}}
        \caption{\textbf{Tukey's Honest Significant Difference (HSD) Test} for each user study criterion and overall global satisfaction. ICC FG (with highest overall satisfaction in Table \ref{tab:userstudy_res}) is in blue.}
        \label{fig:tukey}
    \end{minipage}
    \vspace{-3mm}
    \label{fig:side_by_side}
\end{figure}

\textbf{Results.} Using Friedman's Chi-Square test, we verify that the ranking of the criteria is consistent among the participants ($p < 0.05$). We report normalized criteria importance in Table \ref{tab:userstudy_res} (last column). We observe that Usefulness, Trustworthiness and Actionability are consistently ranked the highest by the participants. We compute the \textit{Global satisfaction} score as a weighted average of the criteria as a global satisfaction measure for each model (Table \ref{tab:userstudy_res}, last row). We additionally conduct pairwise Tukey's Honest Significant Difference (HSD) tests (Fig. \ref{fig:tukey}) to determine the statistical significance of the differences between the means of user preferences per criteria.

We observe that \texttt{ICC FG} ranks the highest in the top 3 most important criteria as well as in global satisfaction. Furthermore, \texttt{ICC GR} ranks second in Usefulness and Actionability and first in Conciseness. Overall, \icc models are favored over interpretable baselines in 4 out of 5 criteria and in terms of global satisfaction. An ANOVA performed separately for each criterion as well as the global satisfaction measure (Table \ref{tab:userstudy_anova}, Appendix \ref{app:userstudy_res}), indicates that there is indeed a significant difference between the models for each criterion. Tukey's HSD tests confirm that \texttt{ICC FG} significantly outperforms NAM and SENN on all criteria but completeness 
(Fig. \ref{fig:tukey}).
\vspace{1mm}
\begin{graybox} Participants prefer \icc explanations in terms of usefulness, trustworthiness, actionability, and conciseness over other intrinsically interpretable models.\end{graybox}

\section{Discussion and Conclusion}
\label{sec:discussion}
\vspace{-3mm}
We proposed \icc, a family of intrinsically interpretable models that puts human understanding at the forefront of the design. Through our experiments on feature gating and group routing (mixture-of-expert) models, we demonstrated that our modular architectures optimize for interpretability but do not compromise on performance. In a real-world setting, we showed that \icc models are preferred over other intrinsically interpretable models in 4 of 5 explainability criteria.

\icc is a locally intrinsic explanation framework that creates explanations that are specific to the input point and guaranteed to be faithful to the model. 
In our architecture, the discriminator network is a black-box model. This is by design, to minimize explanation complexity. We believe the hierarchical prediction logic enabled by \icc is the type of explanation that a user wants: ``Which concepts/features were used to make the prediction, and how important are they to the prediction (weighted sum)?” If the discriminator network was a glass-box model, it would answer the question: ``why were these concepts/features selected for the prediction?”. The first explanation is directly actionable, while the second type of explanation is not. We therefore do not seek to answer this second question with our architecture. However, it would be possible for \icc to have an interpretable discriminator network, as discussed further in Appendix \ref{sec:interpretable-discriminatory}.

\texttt{ICC FG} is best when individual features are important for actionable decisions based on the explanation. However, it can lose sparseness if all features are equally important: consider the extreme case where the prediction is a sum over the full feature space and all features are independent. \texttt{ICC GR} requires more human effort than \texttt{ICC FG} and is, therefore, more suitable for scenarios with cross-feature dependencies and where broader concepts are more actionable than individual features.

\textcolor{black}{\texttt{ICC GR} model's global mixture-of-experts design specializes subnetworks on subsets of features allows them to learn granular patterns (Appendix \ref{app:specialization}). Combining these specialists enhances prediction compared to a monolithic network that might miss underlying patterns (Table \ref{tab:all_results}). Expert-informed feature groupings help the model avoid overfitting to correlations that do not generalize at inference time. For \texttt{ICC FG}, adaptive sparsity (a few features per instance) improves prediction quality by reducing noise and optimizing the interpretability-accuracy tradeoff.}

User-defined feature groups aim at deriving explanations useful to the user, but might compromise performance if the user specifies a grouping that carries minimal signal. Regardless of the grouping, \icc optimizes for explanation actionability and understandability over performance. We believe that an accurate prediction at the cost of explanation usefulness is not worthwhile in an applied setting. \textcolor{black}{It is possible that \texttt{ICC} explanations could be misleading, as concepts used in the explanation could lead to a correlation that was not intended~\citep{zheng2021irrationality, jacovi2021aligning}. In these cases, we view \texttt{ICC} 
 explanations through the lens of auditing model behavior~\citep{yadav2022xaudit}, and encourage human intervention.} \textcolor{black}{\icc's user-centric advantages are highlighted when the input space is human-interpretable. However, for domains that are hard to obtain expert knowledge, we envision \texttt{ICC} increasingly leveraging LLM-extracted features, reducing the necessity of human effort for human-centric explanations~\citep{malberg2024felix, baddour2024phenotypes}.}

We acknowledge that the presentation of explanations in our user study (Section \ref{sec:user-study}) has influence over our preliminary results on the user perception of \icc. We conducted extensive iterations with eight pilot participants to mitigate study design bias. We note that any imbalance in wording is not necessarily in favor of our method; for instance, users found SENN's explanations more complete than \icc (Fig. \ref{fig:tukey}). \textcolor{black}{The user study prioritizes the diversity of study participants and quality of responses over the number of samples evaluated; the task is mentally intensive and we found a longer study can cause a drop in participant attentiveness.} An extensive study over many different tasks and domains of expertise is necessary for generalizable conclusions.

For our text experiments, we fine-tune twenty \texttt{DistilBERT} models as experts (ten for each task). For more complex tasks, for example requiring long context size, multi-step reasoning ability, or strong prior domain-specific knowledge, \texttt{DistilBERT} can be swapped with larger decoder models, either through fine-tuning or in-context learning. Parameter-efficient fine-tuning such as LoRA~\citep{hu2021lora} would allow fine-tuning even large LMs with limited computational cost, while in-context learning would use the same model instance for each feature and feature group. 

Overall, we encourage the machine learning community to design models for interpretability at many different granularities and user-specified requirements. \icc provides one such family of models as a tradeoff between human specification, explanation certainty, and performance.

\section{Acknowledgements}
We kindly thank the Swiss State Secretariat for Education, Research and Innovation (SERI) for supporting this project, and Dr. Tanya Nazaretsky, Bahar Radhmehr, and Paola Mejia Domenzain for helpful discussions. This project has received funding from the European Union’s Horizon 2020 Research and Innovation programme under grant agreement No. 101017915 (DIGIPREDICT).

\bibliographystyle{plainnat}
\bibliography{iclr2025_conference}

\appendix
\section*{Impact Statement} 
This paper introduces work aimed at advancing the field of human-centered machine learning, addressing the critical balance between interpretability and model performance in neural networks. We demonstrate that these aspects do not necessarily need to be a trade-off, particularly in fields where AI can influence decisions that can have effects on people's lives, such as finance, healthcare, and education. 
Focusing on the applications presented in this paper, in healthcare, the ability to understand and trust AI-driven recommendations can greatly improve care delivery. In education, the application of interpretable models reduces the risks of unintended biases, help users to identify actionable interventions, and increases trust.

A key impact of interpretability is regulatory compliance, where guidelines and regulations around AI usage are increasing (e.g. through the EU AI Act). Models like \icc can help industries and organizations meet these legal standards by increasing the transparency of their AI-based decision-making tools.

\texttt{ICC} enhances trust and human-centered actionability in predictive tools by clearly listing the features impacting predictions. This empowers decision-makers to take informed actions. Different contexts and applications have different interpretability needs, such as different levels of granularity, or different sets or actionable concepts. \texttt{ICC} \grouprouting allows users to define their own set of features, adapted to their needs. Moreover, by identifying a finite set of key features, \texttt{ICC} may reduce spurious predictions and allow users to better understand the causes of unexpected model behavior. Overall, an interpretable model like \texttt{ICC} allows users to assess the model's robustness more easily and faithfully on the scale of a individual's decision.

Our tool also offers the potential to identify and mitigate biased behaviors within models, particularly if discriminatory predictions arise from certain feature uses. As different contexts may require varying levels of interpretability, the \grouprouting feature of \texttt{ICC} allows users to define and adjust feature sets according to their specific needs, ensuring flexibility and relevance in diverse applications.

Comparable in societal and technological impact to other interpretable-by-design methods, such as those discussed by \cite{agarwal2021neural}, \texttt{ICC} represents a significant step forward in making AI more accessible and understandable for users and practitioners alike.

We acknowledge that despite the versatility of \texttt{ICC} across various data types, domains, and datasets demonstrated in this work, we cannot guarantee that it will always perform on par with or better than baseline models. It is important to note that the models we enable are not safe for direct use in production tasks, such as detecting breast cancer, without further validation and adaptation by medical professionals. We encourage researchers and practitioners to adopt explainable AI methods.

\section{Taxonomy of Explanation Design Criteria}
\label{sec:related-work}

\textcolor{black}{We include a detailed taxonomy of terms used in Table \ref{tab:approaches}, inspired by \cite{speith2022review,schwalbe2024comprehensive, swamy2023future, pinto2024towards}.}

\noindent \textcolor{black}{\textbf{Granularity:} The granularity (small details or broad concepts) of the explanation \cite{miller2019explanation, swamy2023future, pinto2024towards}.}
\begin{itemize}[leftmargin=1em]
    \item \textcolor{black}{\textbf{Feature:} Explanations are made using the input features.}
    \item \textcolor{black}{ \textbf{Concept:} Explanations are made using concepts (a higher level grouping over the input space).}
\end{itemize}

\noindent \textcolor{black}{\textbf{Basis:} The type of input used by the XAI method/model to create explanations \citep{saeed2023explainable, ghorbani2019towards}.}
\begin{itemize}[leftmargin=1em]
    \item \textcolor{black}{\textbf{Use all input features:} The explanation uses all of the raw inputs to the model.}
    \item \textcolor{black}{\textbf{Concepts:} Either the user defines a concept by specifying examples, the model automatically selects concepts from the feature space, or there is some external definition of concepts grouped over the feature space (user or LLM-specified).}
\end{itemize}

\noindent \textcolor{black}{\textbf{Stage:} The extent to which a model incorporates the explanation in the predictive process \citep{ali2023explainable, speith2022review, schwalbe2024comprehensive}.}
\begin{itemize}[leftmargin=1em]
    \item \textcolor{black}{\textbf{Explain then predict:} The explanation is directly used as part of the prediction process, influencing the model's output.}
    \item \textcolor{black}{\textbf{Explain from model internals:} The explanation reflects the weights or internal mechanisms of the model but is not directly used for prediction (gray box).}
    \item \textcolor{black}{\textbf{Explanation not used in model:} The explanation is completely separate from the model's internals and has no influence on the predictive process (black box).}
\end{itemize}

\noindent \textcolor{black}{\textbf{Sparsity:} The conciseness of the explanation relative to the amount of the input space (features, concepts) used in the prediction \citep{sun2024sparse, ayoobi2023sparx}.}
\begin{itemize}[leftmargin=1em]
    \item \textcolor{black}{\textbf{Sparse:} The explanation uses a minimal amount of features/concepts for prediction.}
    \item \textcolor{black}{\textbf{Not Sparse:} The explanation uses more than a minimal amount of the features/concepts for prediction.}
\end{itemize}

\noindent \textcolor{black}{\textbf{Faithfulness:} Alignment between the explanation and the model behavior, also known as fidelity \citep{lyu2024towards,dasgupta2022framework}.}
\begin{itemize}[leftmargin=1em]
    \item \textcolor{black}{\textbf{Guaranteed:} The explanation aligns to the model behavior with certainty.}
    \item \textcolor{black}{\textbf{Aligned with concepts:} The explanation aligns with the concepts used but is not guaranteed to align with the model behavior.}
    \item \textcolor{black}{\textbf{Approximation:} The explanation is not guaranteed to reflect the model behavior.}
\end{itemize}

\section{Additional Details on Datasets}
\label{sec:appendix-datasets}

Here, we provide additional statistics regarding each dataset used in our study. In particular, we highlight their availability details and terms of use.

\textbf{EDU.} We predict student success during the early weeks of four massive open online courses (MOOCs), using students' clickstream data (see Table \ref{tab:EDU_course_stats} for details about the courses). Contrary to the other 4 datasets, this dataset is kept private for student privacy reasons, as required by HREC 058-2020/10.09.2020 and HREC 096-2020/09.04.2022.

\begin{table}[h!]
\centering
\resizebox{\textwidth}{!}{%
\begin{tabular}{@{}llcccccc@{}}
\toprule
\textbf{Title} & \textbf{Identifier} & \textbf{Topic} & \textbf{Level} & \textbf{Language} & \textbf{No.\ Weeks} & \textbf{No.\ Students} & \textbf{Passing Rate [\%]}  \\ \midrule
Digital Signal Processing & DSP &  CS & MSc & English & 10 & 4,012 & 23.1 \\
Éléments de Géomatique & Geo & Math & BSc & French & 11 & 452 & 45.1 \\
Household Water Treatment and Storage & HWTS &  NS & BSc & French & 5 & 2,438 & 47.2 \\
Villes Africaines & VA &  SS & BSc & En/Fr & 12 & 5,643 & 9.9 \\
\bottomrule
\end{tabular}}
\vspace{2mm}
\caption{Course Details and Statistics.\\
Topic abbreviations: Math: Mathematics; NS: Natural Science; CS: Computer Science; SS: Social Science; Arch: Architecture; Bus: Economics and Business.}
\label{tab:EDU_course_stats}

\end{table}

\textbf{AG News} is a news classification dataset, where given a title and description of a real-world article, it has to be classified into one of the four categories: `World', `Sports', `Business', `Sci/Tech' \citep{zhang2015character}. 
It is freely available at the following location: \url{http://groups.di.unipi.it/~gulli/AG_corpus_of_news_articles.html}, but only for non-commercial use.

\textbf{SST.} The Stanford Sentiment Treebank dataset aims at predicting the sentiment from a sentence fragment sourced from a movie review. The dataset is freely available here: \url{https://huggingface.co/datasets/sst}.
This popular benchmark is an extension of the Movie Review Database (MRD) \citep{socher2013recursive}. It includes two sets of labels: one for binary sentiment classification and one for multiclass. We use binary classification to demonstrate a different setting than the multiclass classification of AG News.

\textbf{Breast Cancer.} The Wisconsin Breast Cancer dataset attempts to identify the presence of cancerous tissue from an image of a fine needle aspirate (FNA) of a breast mass \citep{breastcancer}. This dataset is freely available here: \url{https://archive.ics.uci.edu/dataset/17/breast+cancer+wisconsin+diagnostic}, and is licensed under a Creative Commons Attribution 4.0 International (CC BY 4.0) license.

\textbf{Synthetic Dataset.} We generate 5000 points of OpenXAI's synthetic dataset \citep{agarwal2022openxai} through the class \texttt{generate\_gaussians} with 5 cluster centers, which includes both ground truth labels and ground truth explanations, and is available here: \url{https://github.com/AI4LIFE-GROUP/OpenXAI} (under MIT license).

\section{\icc Group Routing Schema}
\label{sec:appendix-grouprouting}

In the following, we discuss the exact grouping schematic employed on the 4 EDU datasets (including three schemes) and the 2 text datasets.

\textcolor{black}{\textbf{Group Specification Guidance}:  In \icc, we design towards the user's actionability of the resulting explanation. Therefore, if the knowledge that a feature is important can lead to a specific action, and if this action is the same one that should be taken for other features, then those features should be grouped together. From the modeling perspective, grouping features together in a concept means that their shared predictive potential should be leveraged, and likely this is more important for one type of features than another. We maintain that a feature should not be placed in multiple groups to have the faithfulness guarantees that are a strength of \icc. However, in a rare and specific case where two actions must be taken based on the feature, or it is too difficult to decide which concept the feature belongs in, it is always possible to 1) put that feature in its own subnetwork (and have it be selected alongside any of the other feature groups), or 2) combine the two subnetworks into a larger concept.}

\textcolor{black}{Concretely, for \texttt{ICC} GR, users specify a grouping that separates individual features into concepts (which are groups of features). User selected features and user definitions of features are not required in \texttt{ICC} FG. For concept specification, users describe feature groupings in the form of a list or dictionary with which features are assigned to which network. In training, the discriminator takes in all the features as input and decides which of the C concepts to activate. In prediction, there are C predictive modules (one for each concept), each with only the features that are mapped to that concept as input. If a specific concept is activated, then only the features assigned to that concept (by a user) are sent to that subnetwork, and this is used for the eventual prediction.}

\subsection{Feature Grouping Schema for EDU Datasets}\label{app:fg_edu}

The data is composed of student clickstream data from four different MOOCs. The raw clickstream input is transformed into 45 weekly time-series features that have proven useful for student success prediction in previous literature (e.g. total video clicks, forum interactions). 

\textbf{Routing by pattern.} We organize features according to four learning dimensions identified by \cite{asadi2023ripple} and detailed in Table~\ref{tab:mejia_patterns}—effort, consistency, regularity, proactivity, control, and assessment—based on \cite{mejia-patterns}, with a focus on these dimensions in an extended experiment.

\textbf{Routing by Paper.} In routing by paper, we create 4 distinct feature subsets based on handcrafted initial input features from 4 learning science research papers highlighted as predictive for MOOCs in \cite{swamy2022meta}. We filter each feature set to a expert subnetwork for the research paper the set originated from \cite{lalle-conati,marras-vignoud-kaser-2021,chen-cui, boroujeni-et-al}.

\label{gpt-4}
\textbf{Routing by GPT-4. } To leverage LLMs for automated feature groupings, we use the following prompt alongside the feature descriptions from the appendix table of \cite{swamy2022meta}.

\textbf{Prompt}: 

\begin{graybox}
You are a talented learning scientist, specializing in self-regulated learning behavior. You want to group the following features together into smaller groups of logical concepts based on educational theory with headings. What are your feature groupings?

\textbf{Response}: 

Certainly, based on educational theory and self-regulated learning behavior, I would group the provided features into the following logical categories:

\begin{multicols}{2}
\begin{itemize}
    \item Temporal Engagement:
    \begin{itemize}
        \item DelayLecture
        \item RegPeakTimeDayHour
        \item RegPeriodicityDayHour
    \end{itemize}
    \item Study Patterns:
    \begin{itemize}
        \item NumberOfSessions
        \item AvgTimeSessions
        \item TotalTimeSessions
        \item StdTimeSessions
        \item StdTimeBetweenSessions
    \end{itemize}
    \item Video Engagement:
    \begin{itemize}
        \item AvgWatchedWeeklyProp
        \item StdWatchedWeeklyProp
        \item AvgReplayedWeeklyProp
        \item StdReplayedWeeklyProp
        \item AvgInterruptedWeeklyProp
        \item StdInterruptedWeeklyProp
    \end{itemize}
    \item Competency and Learning Progress:
    \begin{itemize}
        \item CompetencyStrength
        \item CompetencyAlignment
        \item CompetencyAnticipation
        \item ContentAlignment
        \item ContentAnticipation
    \end{itemize}
\end{itemize}
\columnbreak
\begin{itemize}
    
    \item Interaction Patterns:
    \begin{itemize}
        \item TotalClicks
        \item TotalClicksProblem
        \item TotalClicksVideo
        \item TotalClicksWeekday
        \item TotalClicksWeekend
        \item TotalTimeProblem
        \item TotalTimeVideo
        \item TotalClicksVideoLoad
        \item TotalClicksVideo
        \item FrequencyEventVideo
        \item FrequencyEventLoad
        \item FrequencyEventPlay
        \item FrequencyEventPause
        \item FrequencyEventStop
        \item FrequencyEventSeekBackward
        \item FrequencyEventSeekForward
        \item FrequencyEventSpeedChange
        \item AvgSeekLength
        \item StdSeekLength
        \item AvgPauseDuration
        \item StdPauseDuration
        \item AvgTimeSpeedingUp
        \item StdTimeSpeedingUp
    \end{itemize}
    \item Assessment Behavior:
    \begin{itemize}
        \item StudentSpeed
        \item StudentShape
    \end{itemize}
\end{itemize}
\end{multicols}
\end{graybox}

\begin{table}[!h]
    \centering
    \begin{minipage}{0.58\textwidth}
        \centering
        \resizebox{\textwidth}{!}{%
        \begin{tabular}{l l l}
            \toprule
           \textbf{Dimensions} & \textbf{Corresponding measures} & \textbf{Student patterns} \\
           \midrule 
           \textbf{Effort} & 
           \begin{tabular}{@{}l@{}}
                Total time online \\
                Total video clicks
           \end{tabular} & 
           \begin{tabular}{@{}l@{}}
                Higher intensity \\
                Lower intensity
           \end{tabular}
            \\
           \midrule
           {\textbf{Consistency}} & 
           \begin{tabular}{@{}l@{}}
               Mean session duration \\
               Relative time online \\
               Relative video clicks
           \end{tabular} & 
           \begin{tabular}{@{}l@{}}
                Uniform \\
                First half \\
                Second half
           \end{tabular}
           \\ 
           \midrule 
           {\textbf{Regularity}} & 
           \begin{tabular}{@{}l@{}}
                Periodicity of week day  \\
                Periodicity of week hour \\
                Periodicity of day hour
           \end{tabular}
            & \begin{tabular}{@{}l@{}}
                 Higher peaks \\
                 Lower peaks 
            \end{tabular} \\
           \midrule 
           {\textbf{Proactivity}} & 
           \begin{tabular}{@{}l@{}}
                Content anticipation \\
                Delay in lecture view
           \end{tabular} & 
           \begin{tabular}{@{}l@{}}
                Anticipated \\
                Delayed
           \end{tabular} \\
           \midrule 
           {\textbf{Control}} & 
           \begin{tabular}{@{}l@{}}
                Fraction time spent (video) \\
                Pause action frequency \\
                Average change rate
           \end{tabular} & 
           \begin{tabular}{@{}l@{}}
               Higher intensity \\
               Lower intensity
           \end{tabular} \\
           \midrule 
           {\textbf{Assessment}} & 
           \begin{tabular}{@{}l@{}}
               Competency strength \\
               Student shape
           \end{tabular} & 
           \begin{tabular}{@{}l@{}}
               Higher intensity \\
               Lower intensity
           \end{tabular}\\
           \bottomrule
        \end{tabular}}
        \vspace{2mm}
        \caption{\textbf{EDU Routing by Pattern} uses learning dimensions from \cite{mejia-patterns} to create interpretable feature groupings.}
        \label{tab:mejia_patterns}
    \end{minipage}
    \hfill
    \begin{minipage}{0.38\textwidth}
        \centering
        \resizebox{\textwidth}{!}{%
        \begin{tabular}{l l}
            \toprule
           \textbf{Code} & \textbf{Field of Study} \\
           \midrule
           \textbf{000} & Computer Science, Information \\
            & and General Works\\
           \textbf{100} & Philosophy and Psychology \\
           \textbf{200} & Religion \\
           \textbf{300} & Social Sciences \\
           \textbf{400} & Language \\
           \textbf{500} & Pure Science \\
           \textbf{600} & Technology \\
           \textbf{700} & Arts and recreation \\
           \textbf{800} & Literature \\
           \textbf{900} & History and geography \\
           \bottomrule
        \end{tabular}}
        \vspace{2mm}
        \caption{\textbf{Text Routing} by the Dewey Decimal Classification system \cite{scott1998dewey}. Each code represents a subnetwork in the text variations of the \icc framework.}
        \label{tab:dewey_decimal_classification}
    \end{minipage}
    \vspace{-5mm}
\end{table}

Each grouping represents a different aspect of self-regulated learning behavior, focusing on how students engage with learning resources, interact with content, demonstrate competency, and approach assessments. This categorization aligns with principles of self-regulated learning and can help in analyzing and understanding students’ behaviors and strategies within an educational context.

\subsection{Feature Grouping Schema for Text Datasets}
\label{app:fg_text}

For news categorization (AG News) and sentiment prediction (SST) feature grouping, we assign words to subnetworks. For this, we use the Dewey Decimal Code (DDC) for librarians and its hierarchy of topics for book classification to create 10 subnetworks, as showcased by topic in Table \ref{tab:dewey_decimal_classification} \citep{satija2013theory}. Each word is assigned to a subcategory (i.e. the word `school' is assigned to the subcategory `education' under category 300 for `social sciences') and routed to the appropriate parent network. The decision to use the DDC was to use subnetworks that were standardized, pervasive in daily life and clearly human-understandable. To conduct this assignment, we utilize SentenceBERT to encode the subtopics for each DDC heading (i.e. all of 010, 020, 030, etc. for the category 000) \citep{reimers2019sentence}.  The choice of SentenceBERT is motivated towards capturing the broader context of multi-word category headings in a lightweight model. During training and inference, we again use SentenceBERT to encode each word in the input instance, then assign each word to the most similar DDC category in embedding space with cosine similarity.

\section{Hyperparameter Sensitivity and Architecture Validation Experiments}
\label{appendix:sparsity}

We examine the sparsity and Gumbel Softmax hyperparameters, and how they impact the \icc model performance. 

\subsection{Sparsity Criterion Experiments}
For the feature gating architecture to further improve interpretability, we would like the network to learn sparse feature activations. That is, for a given input $x$ we would like to reduce the number of features that affect the model prediction. To achieve this we apply regularization to the generated feature mask.

One natural choice to enforce sparse feature activations is to apply $L_1$-norm regularization to the feature mask, which penalizes a high number of nonzero elements. Another choice is to use annealed regularization as presented by Verelst and Tuytelaars \cite{dynconv}, which might aid the model to first work through a prediction optimization phase that is not confounded by any additional error terms before moving towards a sparsity-enforcing phase.

We experiment with annealing L1 and L1 regularization across four courses, and find that traditional L1 regularization is more stable (at least in the time-series setting). The Baseline BiLSTM results are not reported as confidence intervals here as they are directly sourced from a recent benchmarking paper by Swamy et al., with confirmed similarity by Asadi et al. \cite{swamy2022meta, asadi2023ripple}. We reproduce this benchmark above with similar values in Table \ref{tab:sparsity-edu}.

\begin{table}[h]
\centering

\begin{tabular}{@{}llll@{}}
\toprule
\multicolumn{1}{c}{\textbf{\begin{tabular}[c]{@{}c@{}}EDU\\ Dataset\end{tabular}}} &
  \multicolumn{1}{c}{\textbf{Baseline}} &
  \multicolumn{2}{c}{\textbf{\begin{tabular}[c]{@{}c@{}}\icc \\ Feature Gating\end{tabular}}} \\ \midrule
\multicolumn{1}{c}{\textit{40\% EP}} &
  \multicolumn{1}{c}{\textit{BiLSTM}} &
  \multicolumn{1}{c}{\textit{Annealing}} &
  \multicolumn{1}{c}{\textit{L1}} \\ \midrule
DSP       & 82   & 87.76 +/- 3.12 & 90.75 +/- 0.01 \\
Geo       & 76.2 & 81.13 +/- 5.39 & 71.92 +/- 0.01 \\
HWTS      & 72   & 77.58 +/- 0.01 & 82.89 +/- 0.04 \\
VA        & 73.8 & 84.81 +/- 0.01 & 77.80 +/- 0.01 \\ \bottomrule
\end{tabular}
\vspace{2mm}
\caption{\label{tab:sparsity-edu} Annealing L1 regularization in comparison with L1 regularization across EDU datasets. }
\end{table}

\subsection{Gumbel Softmax Hyperparameter Sensitivity} 

We explore the effects of varying the Tau and Threshold parameters during training for \icc FG (top two plots) and GR (bottom two plots) using the DSP course dataset. This dataset was selected due to its extensive use as a benchmark for explainability in education research \cite{swamy2023trusting}. \textcolor{black}{We examine the variability over a tuned discriminator layer (batch size 8, and layer size 32)}, with additional configurations detailed in Appendix \ref{sec:reproduce}. While BAC/ACC remains similar across settings, optimizing the Threshold in conjunction with Tau is crucial for performance; often times changing the threshold is quite a variable experience. Notice the changing impact on BAC across different thresholds as Tau changes.

The thresholds used were 0.5, 0.6, 0.65, 0.7, 0.725, 0.75, 0.775, and 0.8, while the tau values were 0.5, 1.0, 5.0, and 10.0.

\begin{figure}[h]
    \centering
    \includegraphics[width=\linewidth]{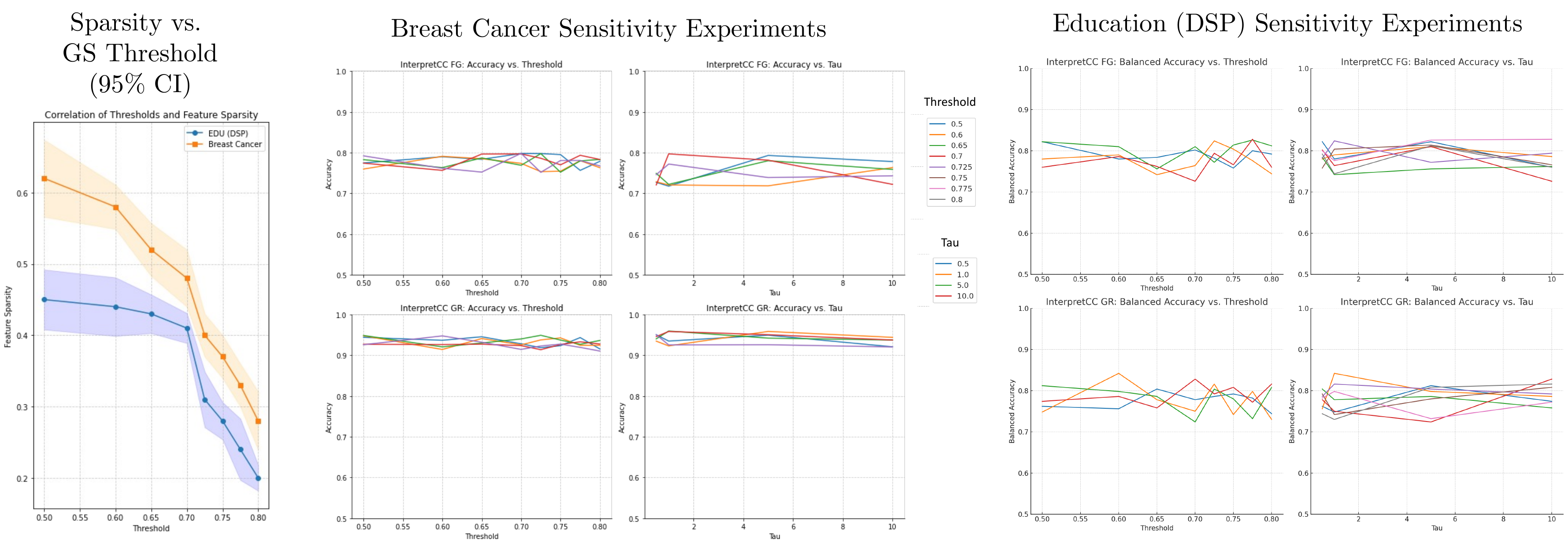}
    \caption{\textbf{\icc Gumbel Softmax Hyperparameter Sensitivity Experiments.} \textcolor{black}{We examine changing Tau and Threshold parameters during training for \icc FG (top two plots) and GR (bottom two plots) for the DSP course dataset and the Breast cancer dataset.}}
    \label{fig:sensitivity}
\end{figure}

\textcolor{black}{All 95\% CIs overlap for the experiments for Figure \ref{fig:sensitivity}. They are omitted from the plots for readability. Significantly high variation (over 0.1) include the following settings:
\begin{itemize}
    \item \icc FG 0.5 threshold, 1.0 Tau with 0.183 standard deviation
    \item \icc FG 0.6 threshold, 5.0 Tau with 0.122 standard deviation
    \item \icc FG 0.65 threshold, 10.0 Tau with 0.151 standard deviation
    \item \icc GR 0.5 threshold, 1.0 Tau with 0.107 standard deviation
    \item \icc GR 0.6 threshold, 5.0 Tau with 0.114 standard deviation
\end{itemize}}

\textcolor{black}{GR architectures are on average 0.043 more stable (less variable) than FG architectures. These experiments show that while the performance of \icc has overlapping 95\% CIs while changing parameters, certain parameter settings have higher variability than others. For both education and health tasks, a $\tau$ of 10 and a Gumbel-Softmax threshold of around 0.7 to 0.8 are performant, sparse in activated features, and relatively stable. Notably, the results for the Breast Cancer dataset are less variable than the DSP dataset, which shows the sensitivity of parameters is domain dependent.}

\subsection{\textcolor{black}{Architecture Validation Analysis}}
\label{app:specialization}

\textcolor{black}{\textbf{“Do we genuinely achieve specialized networks?”} \cite{jacobs1997bias} demonstrates that mixtures-of-experts architectures can optimize the bias-variance trade-off by specializing subnetworks for specific regions of the input space. Similarly, \cite{jiang1999identifiability} prove that the identifiability of mixtures-of-experts models depends on their parameterization.
In the following experiment, we show that \icc  subnetworks specialize to information that are routed to them, and are worse at predicting on information that is not routed to them.}

\begin{figure}
    \centering
    \includegraphics[width=\linewidth]{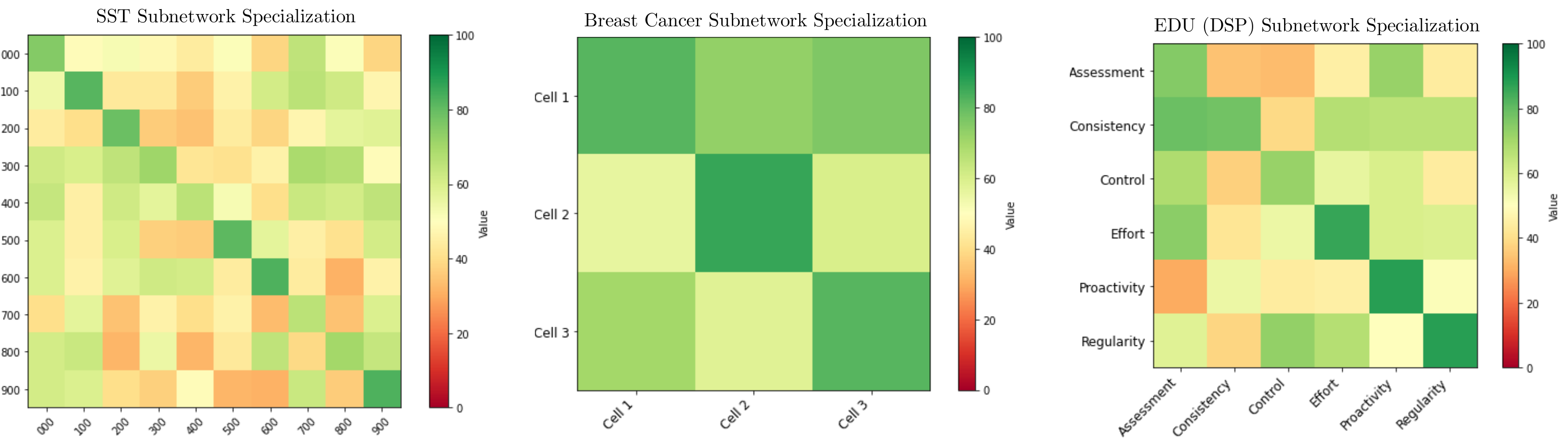}
    \caption{\textcolor{black}{Validation of Specialized Subnetworks: SST, Breast Cancer, and DSP datasets.}}
    \label{fig:heatmaps}
    \vspace{-6mm}
\end{figure}

\textcolor{black}{The experiment, conducted across three different use cases (SST: text, DSP: time series, Breast Cancer: tabular) of \icc, demonstrate that subnetworks predict more strongly in on data that has been routed to them as opposed to data that is supposed to be routed to other subnetworks. Specifically, for DSP, the route-by-pattern networks predict 22.88\% +/- 7.56\% more accurately (balanced accuracy) on points sent to them. For SST, the subnetworks specialize even more strongly (23.59\% +/- 7.86\%). For Breast Cancer, the cell-based grouping predicts 11.88\% +/- 5.57\% better when routed to the same network. The heatmaps in Figure \ref{fig:heatmaps} show the prediction rates of subnetworks (x-axis) on data routed from every other subnetwork (y-axis). Note that across all three use cases, the diagonal performance is the strongest.}

\textcolor{black}{\textbf{“Do these networks contribute to improved prediction accuracy?"} As proven by \cite{jordan1994hierarchical}, mixture-of-experts are a strongly predictive architecture, analogous to the case of ensemble models over monolithic models. The authors show that hierarchical mixtures-of-experts (HMoE) can recursively partition the input space and train effectively using the Expectation-Maximization (EM) algorithm. Empirically, in comparison to post-hoc explainability approaches on top of a non-interpretable base module, we show performance gains (Table 2) for \icc GR in 7 out of 8 datasets, statistically significant higher performance in the Geo and Breast Cancer settings, and comparable performance (overlapping 95\% CIs) in all other settings; this directly shows that using specialized subnetworks is superior to the non-interpretable baseline approach. As \icc FG showcases simply the sparsity requirement with no subnetwork logic, and GR has higher performance than FG results in 4 settings, we know that the subnetworks can be helpful. Similarly, over the non-interpretable base module, FG statistically significantly beats performance in DSP, HWTS, and Synthetic data cases, showcasing the benefits of adaptive sparsity.}

\textcolor{black}{The comparison to NAM showcases the benefits of combining important features together in the predictive module instead of having a separate subnetwork for every feature. The comparison to SENN showcases the benefit of expert-specified routing logic as opposed to automated concept selection. The new comparison to FRESH showcases the benefit of selecting a concept-based explanation over a contiguous explanation. \cite{hazimeh2021dselect} proposes and proves that an architecture with similar foundations in conditional computation achieves efficient and differentiable sparse subnetwork selection, improving task performance and computational efficiency in multi-task learning. They do not focus on an interpretability objective.}

\textcolor{black}{\textbf{“Do these networks contribute to improved interpretability?} Indeed, intrinsic explanations (through hierarchical models or gating) do contribute to increased interpretability \citep{ismail2023interpretable, stojanovic2022bayesian}. From an empirical perspective, the experiments in Table 4 demonstrate \icc’s ability to capture meaningful patterns in the underlying data. The user study addresses whether the outputs of \icc explanations are interpretable to users, and can be useful.}

\section{User Study}
\label{app:user-study}
In this section, we discuss the details of the user study presented in Section \ref{sec:user-study}, discussing the design, content, and additional analyses of the results (including an ANOVA and Tukey HSD tests).
We designed the study over four rounds of pilots, with 8 individuals from diverse backgrounds, continuously updating the study design following their feedback. The survey has been approved by the the Human Research Ethics Committee (HREC) under application number HREC 065-2022/27.09.2022.

\subsection{Design of the Study}

\begin{figure}[h]
    \centering
    \includegraphics[width=\textwidth, trim={5cm 0 25cm 0}, clip]{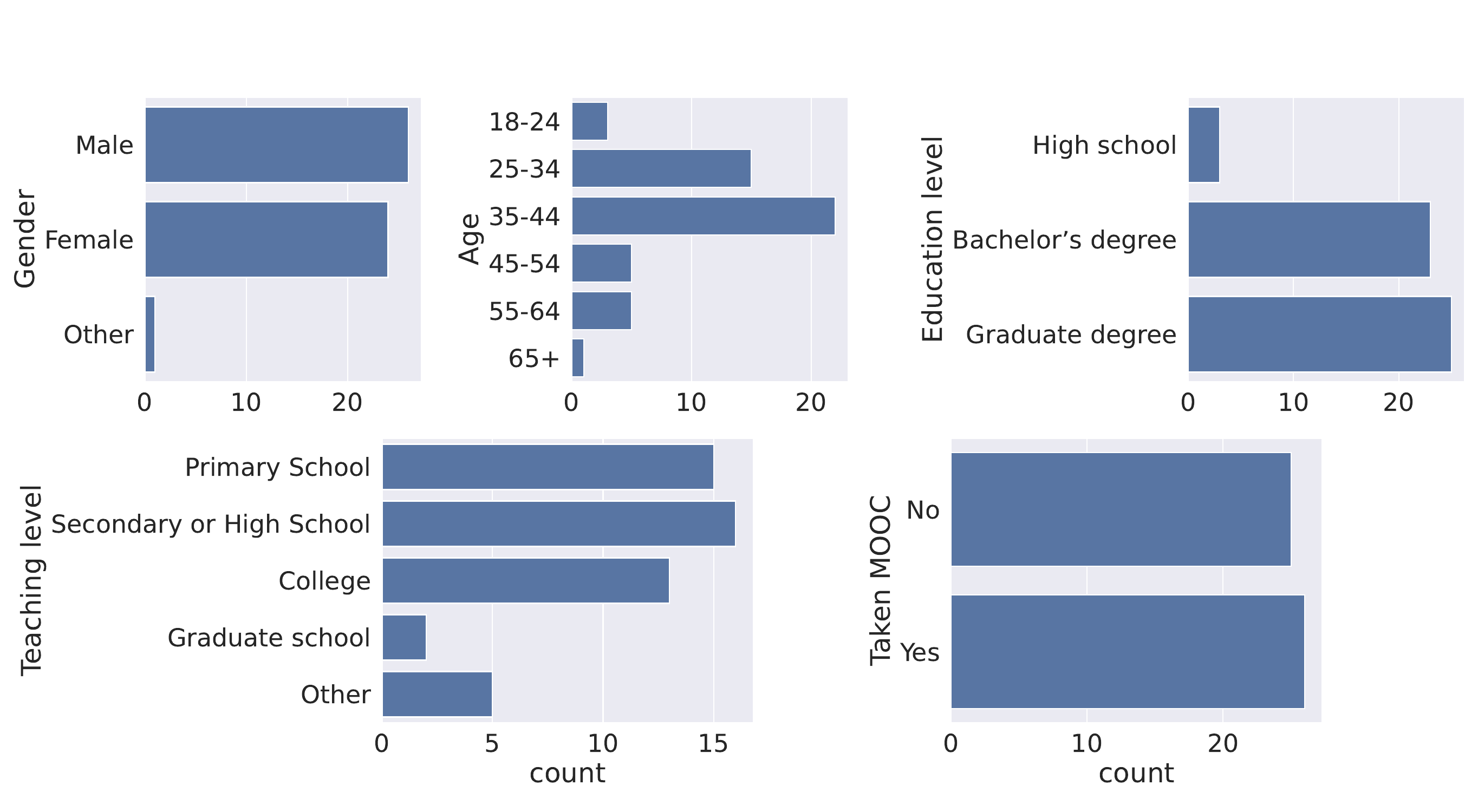}
    \caption{Demographics of teachers that participated in the user study.}
    \label{fig:userstudy_demo}
\end{figure}

We recruit 56 participants using \texttt{Prolific},\footnote{\url{www.prolific.com/}}, selecting the ones who identified their current profession as a teacher and who have at least a bachelor's degree. Our target participants have expertise in teaching, as they would be well-suited to understand both the educational context of the study and the consequences of black-box models for student outcomes. During the study, we ask the participants whether they have ever taken or prepared material for an online course (MOOC), their level of education, and what level they are teaching at (from primary school to graduate school). Detailed demographics distribution can be found in Figure \ref{fig:userstudy_demo}. The sample of participants is gender-balanced, and about half of them have taken or participated in creating a MOOC. The median completion time is 22 minutes, and the average reward per hour is £14.55. 

At the beginning of the study, the participants are shown the following welcome message and disclaimer:\\

\begin{mdframed}[
    backgroundcolor=green!10,
    linecolor=black,
    linewidth=1pt,
    innertopmargin=10pt,
    innerbottommargin=10pt,
    innerrightmargin=10pt,
    innerleftmargin=10pt,
    roundcorner=10pt,
    frametitlebackgroundcolor=gray!30,
    frametitlealignment=\raggedright 
]
Dear participant,

Thank you for participating in our study on model explanations. We are very grateful for your participation and your invaluable insight. Please read this Explanatory Statement in full before moving forward. If you would like further information regarding any aspect of this project, please contact us via the email address provided below.

We are a group of researchers from the [ANON] dedicated to improving education through technology. \textbf{The goal of this study is to evaluate different explanations to help a student improve their performance in an online course.}

- This survey has been approved by the [ANON] Human Research Ethics Committee (HREC) under application number HREC 065-2022/27.09.2022. HREC reviews research proposals involving human participants to ensure that they are ethically acceptable.

 - All the personal information will be kept confidential and anonymized. Only demographic information is being recorded and will only be reported as aggregate in a way that prevents identification of any individual participant. You can freely withdraw at any time and any collected data you provided so far will be destroyed.

 - All data will be collected and stored safely and reported in an anonymous form, in accordance with the [ANON] law on data protection ([ANON]).

- Only anonymized or aggregated data may be used in follow-up research (subject to ethics approval), and made available to other researchers for further analysis and for verification of the conclusions reached by the research team.

- Only the principal investigator and the aforementioned researchers have access to the original data under strict confidentiality. Results from the project may be published in conference papers and/or journal articles. In any case, no personal data will be published (only aggregated, anonymous and/or anonymized data will be published).

- Personal data of participants will be stored for 5 years from the date of collection. During this time, participants have the right to access their data and request information about the processing of their personal data. In order to exercise this right, you need to contact the Principal Investigator.

By participating in this survey, you agree that your data can be used for scientific purposes.

\textbf{In the following study, you will be asked to compare explanations for approximately 35 minutes.} Please ensure that you have enough time to finish the study correctly. Unfinished or only partially answered studies will not be considered as taken part.

We ask you to approach the questions and exercises with seriousness and to complete them to the best of your ability. We will subsequently check questionnaires for seriousness and will have to discard questionnaires that do not meet this requirement.

Thank you for your help. If you encounter any problem with the survey, or if you want to give extra feedback, or receive additional information, feel free to contact us [ANON].
\end{mdframed}

\subsection{Content of the study}

First, we explain the setting of the study to the participants with the following introductory message:

\begin{mdframed}[backgroundcolor=orange!10, linecolor=black, linewidth=1pt, innertopmargin=10pt, innerbottommargin=10pt, innerrightmargin=10pt, innerleftmargin=10pt, roundcorner=10pt]
You are a teaching assistant helping with a Massive Open Online Course (MOOC). This course is taught at the Master's level with quizzes and video lectures taking place over 10 weeks. Since it’s a difficult course with a low passing rate (23.1\%), the teaching team wants to help students who are not doing well to perform better in the course by giving them personalized assistance, and encourage students who are already performing well to continue.

To do this, we have models to predict student success or failure using various weekly behavior features, such as number of video clicks or how accurately questions are answered on the weekly quizzes. If potential failure is predicted early (in our case, in the first 4 weeks of the course), we can use the explanation of the prediction to give additional support (i.e. specific tutoring or assignment reminders) to help the student pass the course.

We train four interpretable machine learning models. Each model predicts a student's performance at the end of the course, in the form of “pass” or “fail”, but also gives us which factors contribute to student success or failure. We want to compare these explanations according to several criteria:
\begin{itemize}[leftmargin=1em]
    \item Usefulness: This explanation is useful to understand the prediction.
    \item Trustworthiness: This explanation lets me judge if I should trust the model.
    \item Actionability: This explanation helps me know how to give feedback to the student.
    \item Completeness: This explanation has sufficient detail to understand why the prediction was made.
    \item Conciseness: Every detail of this explanation is necessary. 

\end{itemize}
\end{mdframed}

We randomly sample 4 students from the test set. Among the 4 selected students, 3 failed the class and one passed. For each student, we predict their success or failure with each model and generate an explanation. We provide them to the participants along with the models' prediction of the student's success or failure. The ground truth (student's performance) and the models' performance are not provided to the participants so that we do not bias their assessment. 

\begin{figure}[ht!]
    \centering
    \includegraphics[width=0.8\textwidth]{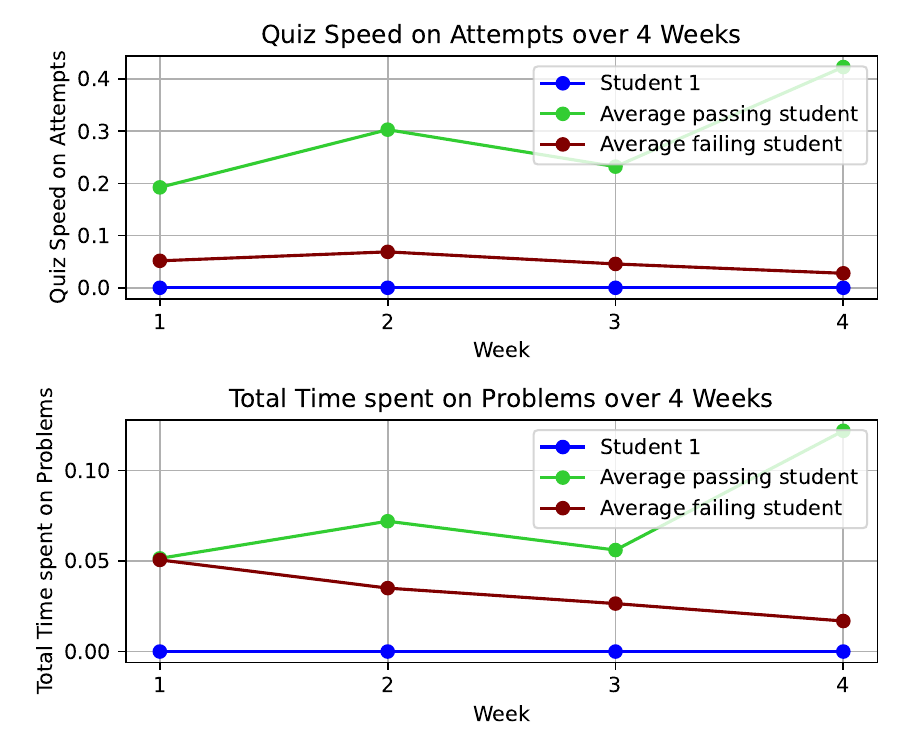}
    \caption{\textcolor{black}{ICC FG method: Importance score visualization of feature-weeks given to the participants for the \icc \featuregating method, for one student.}}
    \label{fig:userstudy-fg}
\end{figure}

\begin{figure}[ht!]
    \centering
    \includegraphics[width=0.8\textwidth]{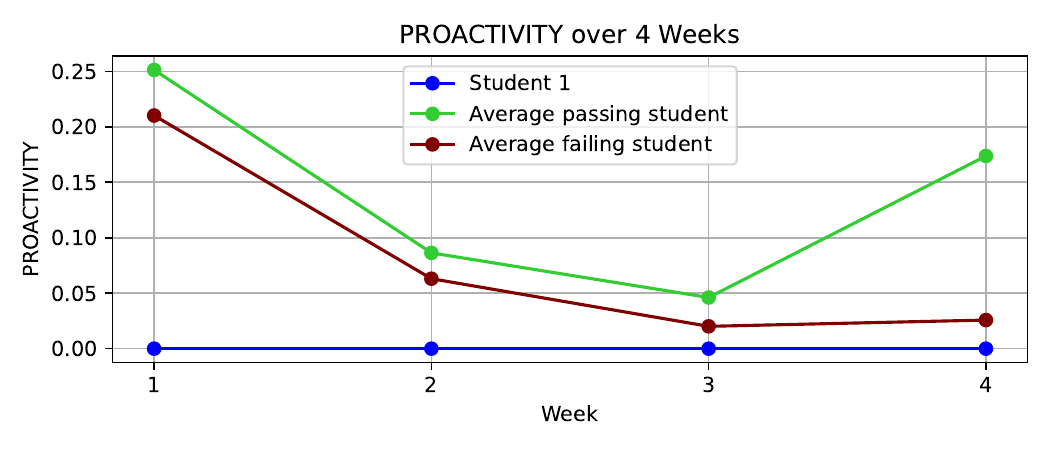}
    \caption{\textcolor{black}{ICC GR method: Importance score visualization of feature-weeks given to the participants for the \icc \grouprouting with pattern-based matching method, for one student.}
}
    \label{fig:userstudy-pr}
\end{figure}

\begin{figure}[ht!]
    \centering
    \includegraphics[width=\textwidth]{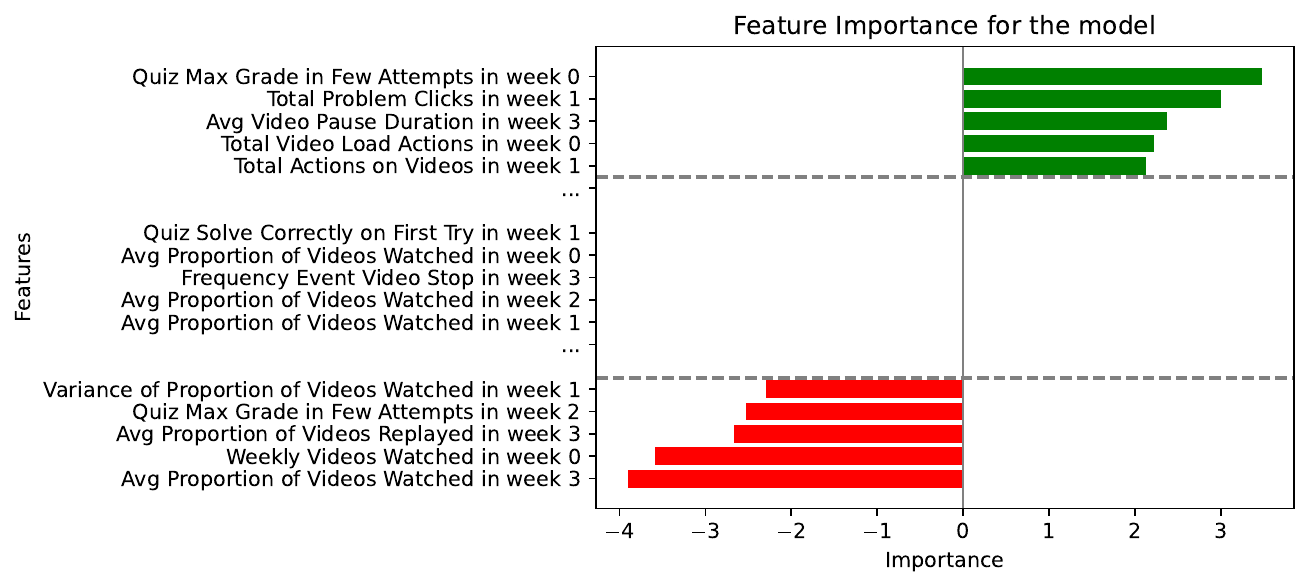}
    \caption{\textcolor{black}{NAM method: Importance score visualization of feature-weeks given to the participants, for one student}.
}
    \label{fig:userstudy-nam}
\end{figure}

\begin{figure}[ht!]
    \centering
    \includegraphics[width=\textwidth]{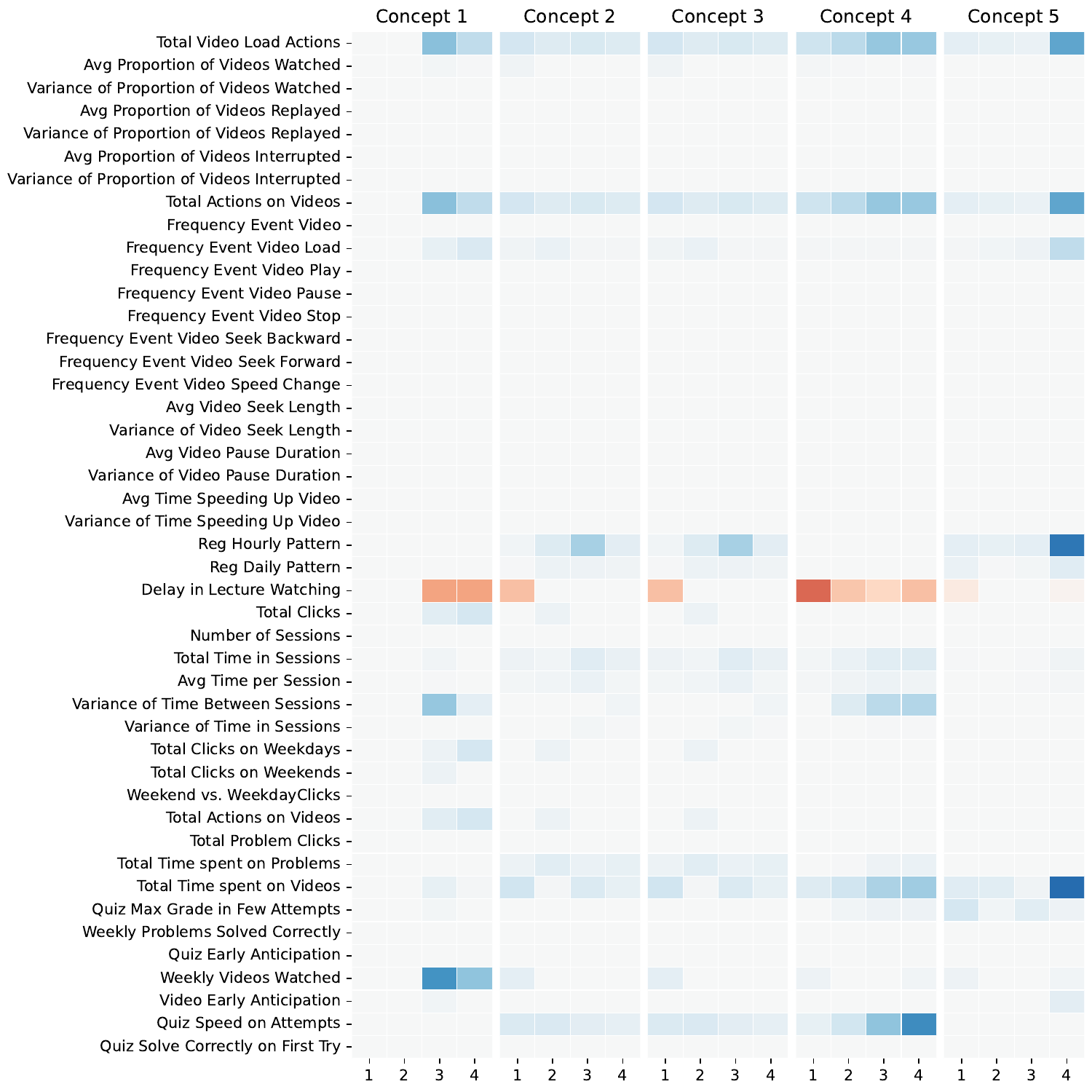}
    \caption{\textcolor{black}{SENN Method: Importance score visualization of feature-week given to the participants, for one student}.}
    \label{fig:userstudy-senn}
\end{figure}

The content of the explanation obtained by each method differs greatly. We simplify the explanations and render them in textual and graph format to make them as easy to understand as possible to a human.  For \icc explanations, we provide the full list of single features / feature groups used by the models. For \featuregating, we show the evolution of the student's behavior across weeks for each of the features used (see Figure \ref{fig:userstudy-fg}). For \grouprouting, we compute a generic score for each concept used by the model, by averaging the normalized behavior features that compose the concept. We show the evolution of that concept measure across weeks for the student. We also provide the definition of the concept and of the features that compose it (see Figure \ref{fig:userstudy-pr}). For SENN, we select the top 5 groups of students, that we call \textit{concepts}. We showcase all 180 feature-weeks for each concept along with their importance in that concept, highlighting the salient ones. We also provide the importance of each concept for the model's prediction (see Figure \ref{fig:userstudy-senn}). Finally, NAM assigns feature importance to all 180 feature-weeks used to make the prediction. We select the 5 feature-weeks found to have highest positive impact, 5 feature weeks with the lowest impact, and the 5 feature-weeks with highest negative impact. We show their importance in a barplot (see Figure \ref{fig:userstudy-nam}).

Note that the choice we made for the presentation of the explanations might have an influence on the participants' perception of the explanations. In an ideal setting, we would provide a very detailed description of how each model uses the features and how the explanation is derived, so that the participants can fairly assess the explanation's quality. However, in a realistic setting, the user facing the explanation might not have the time or prior knowledge necessary to understand these elements. To tackle that trade-off between ensuring thoroughness and accessibility of the content of each model's explanation, we opted for a balanced approach in presenting each explanation with a simple graph and an explanatory text that we kept as short as possible.

\begin{figure}[ht!]
    \centering
    \includegraphics[width=\textwidth]{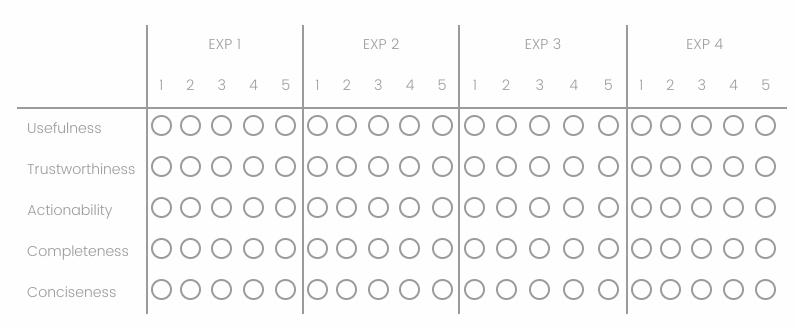}
    \caption{Format of the Likert scale question asked for each criterion and explanation.}
    \label{fig:userstudy-likertscale}
\end{figure}

For each new sample (student taking the course), we provide the list of 4 model explanations in random order. We ask participants to compare these explanations according to five criteria using likert scales. A screenshot of the answer section is shown in Figure \ref{fig:userstudy-likertscale}. We include a practice question to train the participants in how to answer the study and filter inattentive experts. We excluded from the analysis 5 participants who failed to answer correctly to half of the practice questions.

Along with the \icc FG graphs, we provide the following explanatory text:
\begin{graybox}
This student is predicted to fail the course. \textbf{The model found the following 2 features to be the most predictive for this student, and only used these features to make the prediction}:
\begin{itemize}[noitemsep,left=0pt]
\item Quiz Speed on Attempts: The average time passed between two consecutive attempts for the same quiz.
\item Total Time spent on Problems: The total (cumulative) time that a student has spent on problem events.
\end{itemize}
The plot shows the evolution of the student's behavior for each feature across the 4 weeks.
\end{graybox}

Along with the \icc GR graph, we provide the following explanatory text:
\begin{graybox}
This student is predicted to fail the course. \textbf{For this student, the model decided to only use 1 group(s) of features to make the prediction as this was the one(s) it found most important. The groups of features were designed by expert literature in learning sciences}:
\begin{itemize}[noitemsep,left=0pt]
\item Concept: PROACTIVITY
\end{itemize}
Definition: Proactivity measures the extent to which students are on time or ahead of the schedule, as engagement in pre-class activities has shown to be associated with exam performance.
\begin{itemize}[noitemsep,left=0pt]
\item It includes the following features: The number of videos covered by the student from those that are in subsequent weeks and The average delay in viewing video lectures after they are released to students.
\end{itemize}
\end{graybox}

Along with the NAM graph, we provide the following explanatory text:
\begin{graybox}
This student is predicted to fail the course. \textbf{The model used all 180 feature-weeks (45 features from 4 weeks) to make the prediction. It has assigned a level of importance for each feature-week, showing how much it impacts the predictions, independently of the student's behavior.}
Out of the 180 feature-weeks, the plot shows the 5 feature-weeks found to have the highest positive impact, 5 feature weeks with the lowest impact, and the 5 feature-weeks with the highest negative impact. For example, Quiz Max Grade in Few Attempts in week 0 has an importance score of 3.48.
\end{graybox}

Along with the SENN graph, we provide the following explanatory text:
\begin{graybox}
This student is predicted to fail the course. \textbf{The model used all 180 feature-weeks (45 features from 4 weeks) to make the prediction. It groups them into 5 concepts automatically and assigned a score to each concept.} Each concept can be interpreted as a group of features that are important for the prediction. The plot shows the importance of each feature-week for each concept, from high positive importance (blue) to high negative importance (red).

Here is the list ordered by absolute value of importance:
\begin{itemize}[noitemsep,left=0pt]
    \item Concept 3 (-0.21)
    \item Concept 2 (0.08)
    \item Concept 1 (-0.01)
    \item Concept 5 (-0.01)
    \item Concept 4 (0.01)
\end{itemize}
\end{graybox}

\subsection{Significance Testing for User Study Results}\label{app:userstudy_res}

We perform an ANOVA to determine the effect of the model and the sample (the student) on the score given by the participants, for each criterion and on average.
Table \ref{tab:userstudy_anova} shows the p-values testing the significance of the effect of the model and sample on the participants' scores for each criterion. It can be interpreted the following way. In the first row, if the p-value is lower than the significance level (0.05), then there is a statistically significant difference in scores across the different models for that criterion. In the second row, if the p-value is below 0.05, there is a statistically significant difference in scores across the different students. Finally, a p-value lower than 0.05 in the final row shows that the effect of the model on the scores depends on the student. 
According to the table, the model has always a significant impact of the value given to each criterion. 
Then, we apply Tukey's Honest Significant Difference (HSD) Test to determine, for each pair of explanation, if their scores are significantly different (Figure \ref{fig:tukey}).

\begin{table}[h]
    \centering
    \resizebox{\textwidth}{!}{%
\begin{tabular}{lrrrrrr}
\toprule
{} &  Usefulness &  Trustworthiness &  Actionability &  Completeness &  Conciseness &  Global \\
\midrule
Model         &       0.000 &            0.001 &          0.000 &         0.000 &        0.000 &   0.000 \\
Student       &       0.004 &            0.018 &          0.144 &         0.003 &        0.075 &   0.075 \\
Model:Student &       0.143 &            0.058 &          0.063 &         0.000 &        0.171 &   0.171 \\
\bottomrule
\end{tabular}}
    \caption{ANOVA results for the user study.}
    \label{tab:userstudy_anova}
\end{table}

Extended description for Figure \ref{fig:tukey} -- Significance test using Tukey's Honest Significant Difference (HSD) Test. It indicates which pairs of models have significantly different means. We highlight the top model on average (\textit{Global satisfaction}, in the last plot), ICC FG (\icc \featuregating) in blue, and the models that are significantly worse according to each criterion in red. Example of interpretation using the Tukey HSD test: for the criterion \textit{Usefulness}, ICC FG is scored significantly better than SENN and NAM, but is not significantly better than ICC GR.

\section{Reproducibility and Compute Details}
\label{sec:reproduce}
All EDU, Health, and Synthetic experiments were run on a single NVIDIA A100 GPU with 32 GB Memory, each model taking approximately 20 minutes (or often less) to train. Text experiments for InterpretCC took approximately an hour to train. SENN Concepts took the longest time of all models, with each model running within 3-4 hours.

All interpretable models reported in \ref{tab:all_results} and \ref{tab:group-routing} has been hyperparameter tuned over the following parameters with early-stopping (where applicable): 
\begin{itemize}
    \item learning rate: 1e-3, 1e-4, 2e-5, 1e-5
    \item layer size: 16, 32, 64
    \item number of concepts (only for SENN): 5, 6, 7
    \item batch size: 8, 16, 32, 64)
    \item gumbel softmax threshold (only for \icc): 0.1, 0.3, 0.5, 0.7
\end{itemize}
Each Feature-Based model was run for 100 epochs with early stopping, and each Concept-Based model was run for 20 epochs with early stopping. Other details related to preprocessing and thresholds are included directly in our repository.

\section{Gumbel Softmax trick and its application to \icc}
\label{appendix:gumbel}
To make the feature gating and routing architectures compatible with backpropagation, we need to make the masks differentiable. These discrete decisions can be trained end-to-end using the Gumbel Softmax trick \cite{gumbel}. This method adapts soft decisions into hard ones while enabling backpropagation, i.e. provides a simple way to draw samples from a categorical distribution. 

Given a categorical distribution with class probabilities $\pi = [\pi_1 \, \pi_2 \, ... \, \pi_N]$, one can draw discrete samples $z$ as follows: 
$$z = \textsc{OneHot}\left(\arg\max_i [g_i + \log \pi_i]\right)$$

where $g_1 ... g_N$ are i.i.d. samples drawn from the Gumbel$(0, 1)$ distribution. Then, the softmax function is used as a differentiable approximation to $\arg\max$ to generate a $N$-dimensional sample vector $y$ such that
$$y_i = \frac{\exp((\log(\pi_i) + g_i)/\tau)}{\sum_{j=1}^N\exp((\log(\pi_j) + g_j)/\tau)} \quad \text{for } i = 1, ..., N$$
where $\tau$ is a softmax temperature parameter that is fixed at $\tau = 1$ for experiments in this project. 

Notice that for the gating mechanism, an independent sample is drawn for each `gate' instead of for each datapoint in routing. For example in feature gating, for each feature $i$, a soft-decision $a_i \in (-\infty, +\infty)$ is outputted by the discriminator layers. The probability $\pi_1$ that the feature should be activated as well as the complement probability $\pi_2$ (feature is not activated) can then be computed by using the sigmoid function:
$$\pi_1 = \sigma(a_i) \qquad \pi_2 = 1 - \pi_1 = 1 - \sigma(a_i)$$ The corresponding (1-dimensional) sample $y$ for each $i$ can thus be reduced to
$$y = \sigma\left(\frac{a_i + g_1 - g_2}{\tau}\right)$$
In other words, the discriminator layers from Fig. \ref{fig:arch} actually feed into an adapted Gumbel Sigmoid where $\sigma_i$ is the corresponding $y$ sample as described above.

For routing, the discriminator layers actually output the route logits to a Gumbel Softmax, which constructs the categorical sample vector (of dimension equal to the number of routes and $i$-th entry $y_i$ defined as above).

Finally, we can use a straight-through estimator during training. In other words, binary (or hard/quantized) samples are then used for the forward pass while gradients are obtained from the soft samples for backpropagation. This means that, given soft decisions $\sigma_i$,  architectures that use a mask $M = [m_1 \, ... \, m_N]$ with $m_i = \mathds{1}_{\{\sigma_i \geq 0.5\}}$ differ in value during the forward and backward pass:
$$m_i = \begin{cases}
    \mathds{1}_{\{\sigma_i \geq 0.5\}} \qquad \text{(forward pass)}, \\
    \sigma_i \qquad\qquad\quad \text{(backward pass)}
\end{cases}$$

\section{Additional Group Routing Experiments}\label{app:group_routing}

We conduct additional analyses for the high impact, real-world applications in Health, Text, and EDU.

\subsection{Grouping Analysis for Breast Cancer Dataset} 

\begin{figure}
     \begin{minipage}[b]{0.48\textwidth}
     \centering
     \includegraphics[width=\linewidth, trim={5 5 5 5}, clip]{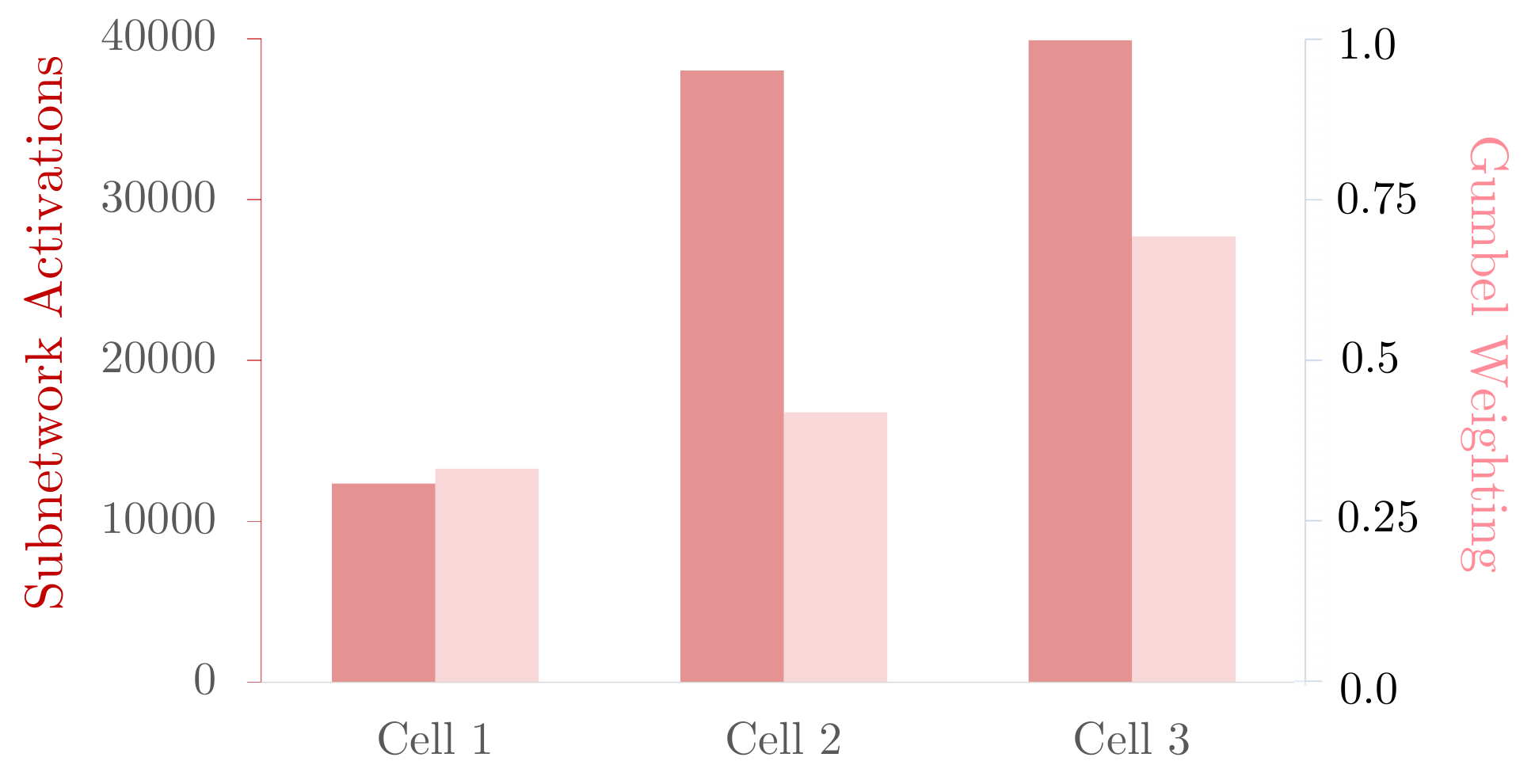}
        \caption{\textbf{Breast Cancer}: Number of activations (left) vs. average activation weight (right) per cell. Each subnetwork represents a distinct cell nuclei (10 features).}
        \label{fig:cancer-comparison}
    \end{minipage}
\hfill
\begin{minipage}[b]{0.48\textwidth}
        \centering
        \includegraphics[width=\linewidth, trim={5 5 5 5}, clip]{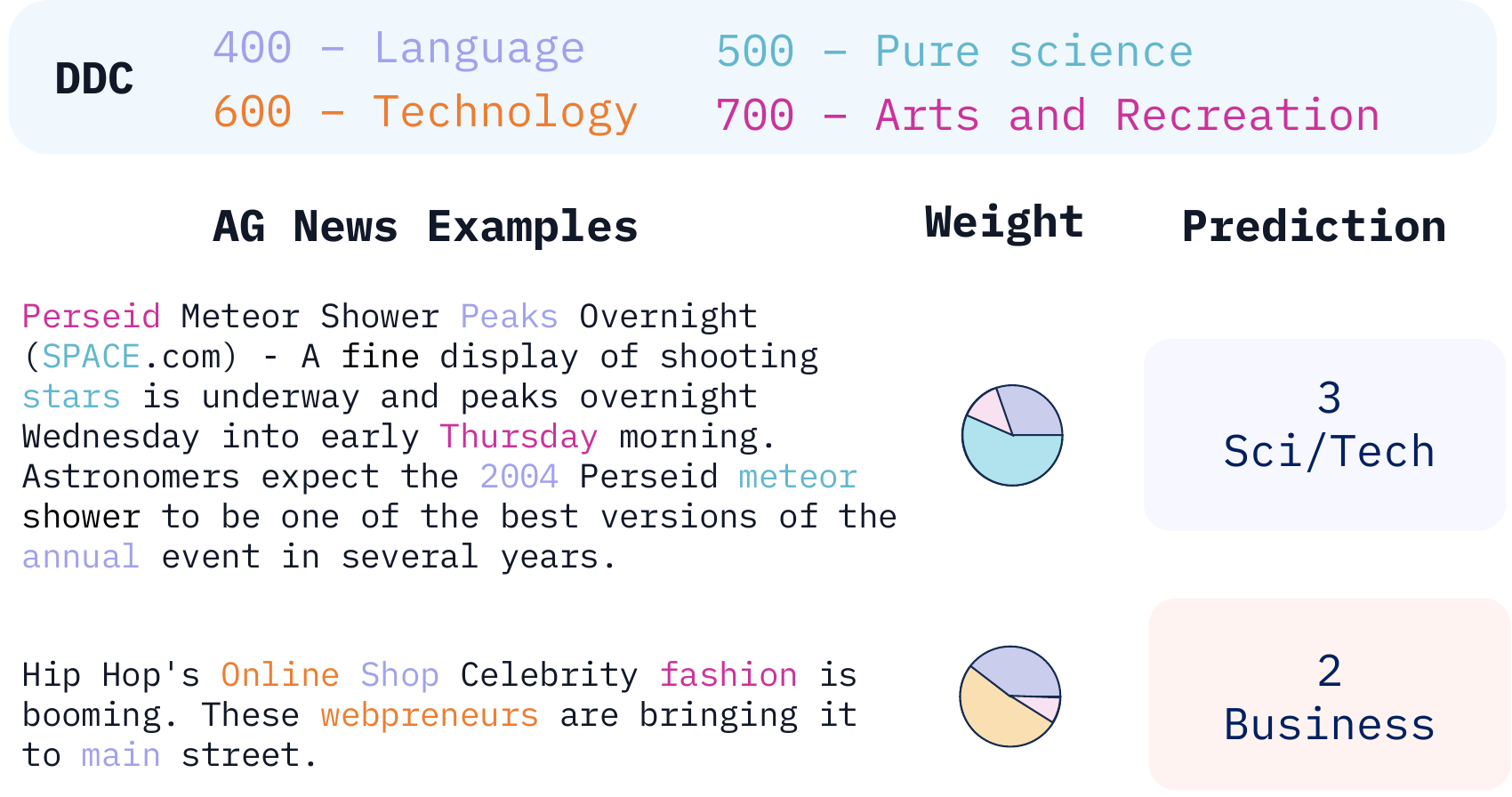}
        \caption{\textbf{AG News}: Two qualitative examples from \textbf{AG News} with \texttt{ICC} explanations, where highlighted text represents a subnetwork activation. The pie chart represents the weight of each subnetwork in the resulting prediction. }
        \label{fig:text-example}
    \end{minipage}
\end{figure}

For the \textbf{Breast Cancer} data set, the subnetworks grouping features from Cell $1$ and Cell $2$ are activated much more frequently than the third subnetwork (see Fig. \ref{fig:cancer-comparison}). Furthermore, Cell $3$ also gets activated with higher weights than the other two cells (Cell $1$: $0.25$, Cell $2$: $0.40$, Cell $3$: $0.0.70$). Smoothness and texture of the tissue images were the most important features across cells.

\subsection{Grouping Analysis for Text Datasets}
We provide two illustrative, qualitative examples from the AG News dataset in Figure \ref{fig:text-example}. In the top example (Perseid meteor shower), the words `stars', `meteor', and `SPACE' are routed to the \textit{Pure Science} (500) subnetwork with a $50\%$ activation weight, resulting in the correct prediction of `Sci/Tech' category. Likewise, for the bottom article, both the \textit{Technology} and \textit{Arts} subnetworks are highly weighted, resulting in the correct prediction of the `Business' category. Interestingly, subnetwork \textit{Language} (400) is also activated.

\subsection{Grouping Analysis for EDU Datasets}
\label{appendix:edu}

We conduct sparsity and group routing paradigm analyses on a representative course of the EDU datasets (DSP). We additionally experiment with grouping by paper and pattern over all courses, including two additional MOOC courses with low predictive performance (Structures and Ventures) featured in \cite{swamy2022meta}.

\subsubsection{Sparsity Analysis on DSP}\label{appendix:edu_dsp_sparsity}
We conduct a sparsity analysis on the course used for the user study, DSP, and featured in several research works in the learning science community \cite{swamy-eval, swamy2022meta, boroujeni-et-al}. In line with \cite{swamy2022meta}, we evaluate the sparsity at both the 40\% and 60\% prediction setting, representing early prediction at 4 and 6 weeks of the course respectively.

\begin{table}[h]
    \centering
    \begin{tabular}{c c c c}
        Setting & Model & 
        \begin{tabular}{c}
             Performance\\
             \begin{tabular}{|c|c|c|}
                \hline
                 $\;$ACC$\;$ & $\,$BAC$\,$ & AUC$\;$ \\
                 \hline
             \end{tabular}
        \end{tabular} & 
        \begin{tabular}{c}
             Activated Features\\
             \begin{tabular}{|c|c|}
                \hline
                 $\;\;\;\mu\;\;\;$ & $\,\,\;\sigma\;\,\,$ \\
                 \hline
             \end{tabular}
        \end{tabular}\\
        \\
        40\% & 
        \begin{tabular}{c}
            Baseline \\
            L1 (1e-5)\\
            L1 (1e-4)\\
            Annealed MSE \\
            Truncated AMSE \\
        \end{tabular} &
        \begin{tabular}{|c|c|c|}
            \hline
            0.835 & 0.653 & 0.858 \\ 
            0.790 & 0.711 & 0.817 \\ 
            0.763 & 0.748 & 0.784 \\ 
            0.768 & 0.770 & 0.823 \\ 
            0.805 & 0.743 & 0.823 \\
            \hline
        \end{tabular} & 
        \begin{tabular}{|c|c|}
            \hline
            97 & 0 \\ 
            8.54 & 0.58 \\ 
            2.20 & 1.64 \\ 
            13.26 & 3.72 \\
            5.95 & 1.03 \\
            \hline
        \end{tabular}
        \\
        60\% & 
        \begin{tabular}{c}
            \\
            Baseline \\
            L1 (1e-5)\\
            L1 (1e-4)\\
            Annealed L1 (1e-5)\\
            Annealed MSE \\
            Truncated AMSE \\
        \end{tabular} &
        \begin{tabular}{c}
            \\
            \begin{tabular}{|c|c|c|}
                \hline
                0.944 & 0.925 & 0.982 \\
                0.914 & 0.935 & 0.963 \\ 
                0.914 & 0.917 & 0.957 \\ 
                0.910 & 0.927 & 0.957 \\
                0.892 & 0.927 & 0.952 \\
                0.787 & 0.788 & 0.839 \\  
                \hline
            \end{tabular}  
        \end{tabular} & 
        \begin{tabular}{c}
            \\
             \begin{tabular}{|c|c|}
                \hline
                97 & 0 \\
                37.60 & 3.97 \\ 
                31.38 & 4.45 \\ 
                38.36 & 3.61 \\
                21.45 & 3.23 \\
                - & - \\
                \hline
            \end{tabular} 
        \end{tabular}
    \end{tabular}
    \vspace{2mm}
    \caption{\textbf{\icc Feature Gating} comparison of performance metrics between different sparsification methods next to their average and standard deviation of number of activated features for both 40\% and 60\% early success prediction settings; baseline benchmarks also provided for contrast.}
    \label{tab:fg_results}
\end{table}

An annealed mean-squared regularization proved most effective, although it activated more features on average than $L_1$-norm regularization which more effectively reduced the feature space while achieving desirable balanced accuracy performance in this setting. By truncating the initial feature space to only the activated features and using the same architecture, performance is almost maintained even though the average number of activated features per datapoint is more than halved (from around 13 to 6). However, this method was not as effective for the 60\% setting. Truncating the feature space largely reduces predictive capability (e.g. almost a 0.15 drop in balanced accuracy). L1 regularization in this case proved best. Using an annealed regularization did not significantly improve or change model performance as well. 

\subsubsection{Multiple Grouping Paradigms on DSP}\label{appendix:edu_dsp_gr}

\begin{figure}
    \centering
    \includegraphics[width=\linewidth, trim={5 5 5 5}, clip]{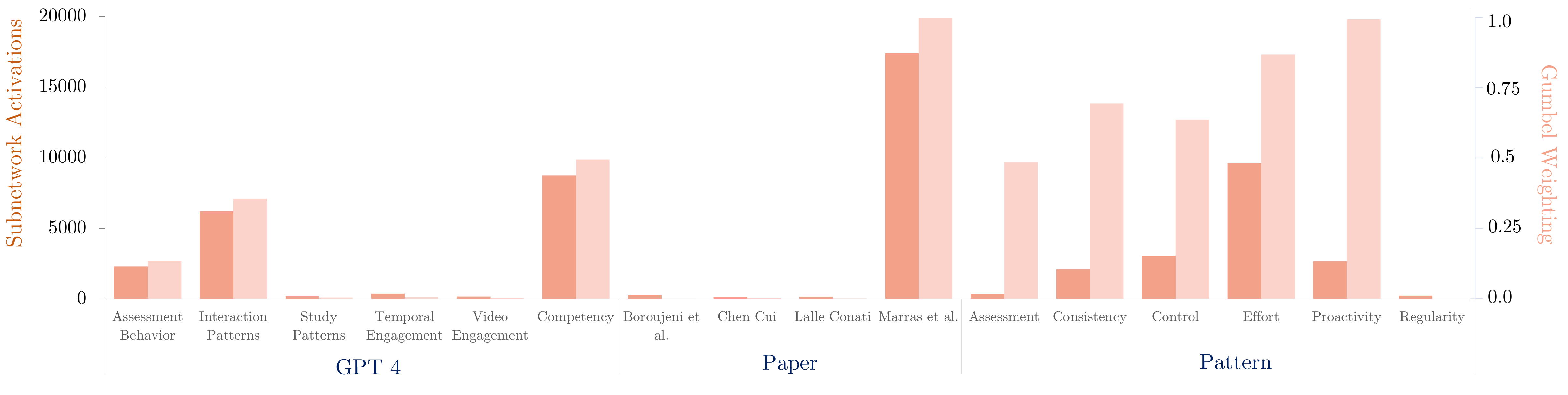}
    \caption{\textbf{EDU}: Number of subnetwork activations  (left) and Gumbel Softmax activation weights (right) across different groupings (GPT-4, Paper, Pattern) for course DSP 1 of the EDU domain.}
\label{fig:deep-dive}
\end{figure}

To illustrate the influence of different feature groupings, we conduct a deep dive for course DSP 1 of the EDU domain. Figure \ref{fig:deep-dive} illustrates the number of subnetwork activations and corresponding weights for three different groupings.

For the first two groupings (GPT-4, Paper), the subnetwork activations (number of times the route was activated) closely mirror the Gumbel Sigmoid adaptive weighting, indicating that a few networks are activated with high weights for prediction. In the \textit{group by GPT-4 }setting, we see behaviors of competency, interaction patterns, and assessment frequently activated for student pass-fail predictions. Although `interaction patterns' is the largest category (most number of features chosen by GPT-4), it still comes second to competency (focusing on student achievement). In the \textit{group by paper} setting, we see a clear preference for Marras et al. with over $17,500$ students predicted using this network (dark orange) and high weight given to the predictions from the network (light orange).

In contrast, in the third grouping (Pattern), we see a differentiation between the number of activations (dark orange) and the weight of the activations (light orange). Notably, the patterns of `Effort', `Proactivity', `Consistency' and `Control' all have higher than $50\%$ weight when they are activated, which means they contribute a lot to the overall prediction when chosen.

\subsubsection{Grouping by paper and pattern over all courses}\label{app:fg_allcourses}

\begin{figure}[h]
    \centering
    \includegraphics[width=0.6\linewidth, trim={5 5 5 5}, clip]{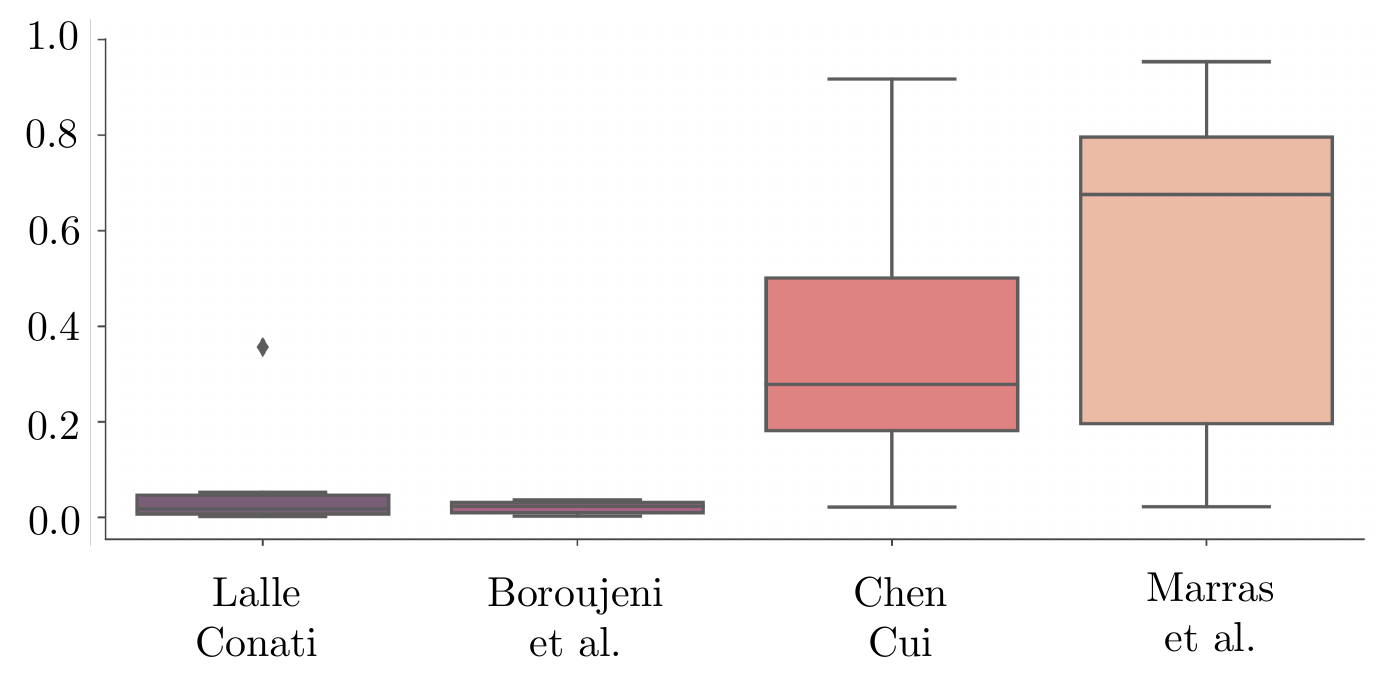}
    \caption{\textbf{EDU} analysis of \textit{group by paper} routing averaged over all six courses for each paper grouping. The y-axis represents the proportion of points for which the subnetwork is activated.}
    \label{fig:edu-group-by-paper}
\end{figure}

In Figure \ref{fig:edu-group-by-paper}, we see \icc routing by research paper (grouping the features based on the paper they were proposed in). The Marras et al. and Chen Cui feature sets have clearly been identified as important over the majority of courses, echoing findings in other learning science literature using BiLSTM and random forest architectures \cite{marras-vignoud-kaser-2021, chen-cui, swamy2023trusting}. The large standard deviations in the box-plots indicate that for at least some courses (in this case Structures and Venture), Chen Cui and Marras were not found significantly important. Notably, the same courses that have low accuracies on routing in \ref{tab:all_results} are those that have low scores on the two most popular feature sets, showing a consensus among performant \icc models and a validation of the identification of importance. 

\begin{figure*}[h]
    \centering
    \includegraphics[width=\textwidth, trim={5 5 5 5}, clip]{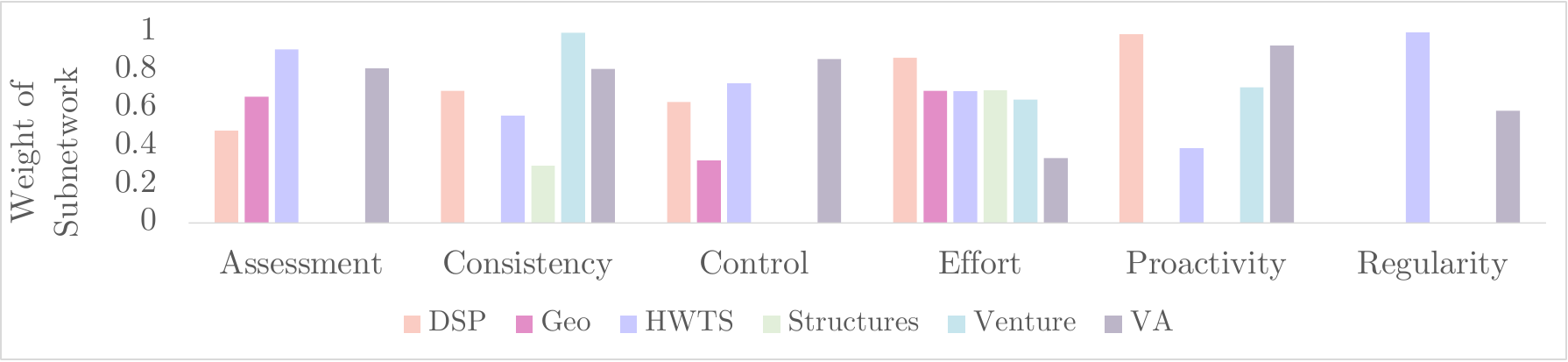}
    \caption{\textbf{EDU} Weighting distribution across subnetworks using \textit{group by pattern} for 6 courses.}
    \label{fig:edu-network-weights}
\end{figure*}

\begin{figure}[h]
    \begin{minipage}[]{0.55\textwidth}
    \centering
    \resizebox{\textwidth}{!}{
        \begin{tabular}{@{}lll@{}}
        \toprule
        \multicolumn{1}{c}{\textbf{Model}} & \multicolumn{1}{c}{\textbf{B. Cancer}} & \multicolumn{1}{c}{\textbf{Synthetic}} \\ \midrule
        Random Forest (RF) & \textit{92.98 ± 4.70} & 85.32 ± 3.48 \\ \midrule
        \begin{tabular}[c]{@{}l@{}}Non-interpretable \\ base module (NN)\\ (SHAP, LIME)\end{tabular} & 89.70 ± 1.05 & 86.67 ± 0.31 \\ \midrule
        NAM & 88.77 ± 7.31 & 87.85 ± 1.31 \\ \midrule
        SENN Features & 80.52 ± 6.21 & 83.67 ± 1.86 \\ \midrule
        SENN Concepts & 85.26 ± 1.03 & \textit{89.51 ± 0.51 }\\ \midrule
        InterpretCC FG & 78.19 ± 3.54 & 84.67 ± 4.04 \\ \midrule
        InterpretCC Top K & 84.66 ± 3.02 & \textbf{90.83 ± 1.93} \\ \midrule
        InterpretCC GR & \textbf{94.85 ± 1.25} & 89.47 ± 2.89 \\ \bottomrule
        \end{tabular}}
    \end{minipage}%
    \hfill
    \begin{minipage}[]{0.4\textwidth}
        \centering
        \includegraphics[width=\linewidth]{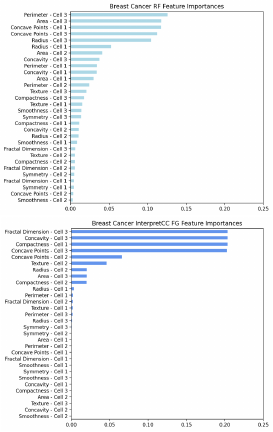}
    \end{minipage}
    \caption{\label{fig:icc-vs-rf}\textbf{\icc vs. RF Tabular Comparison}. (left) Performance of models on Breast Cancer and Synthetic datasets. (right) Features chosen in explanations from RF (light blue) vs. \icc for Breast Cancer dataset (dark blue).}
\end{figure}

In Fig. \ref{fig:edu-network-weights}, we see a widely varying distribution of patterns selected across courses, showcasing the ability of \icc to adaptively select subnetwork weights depending on the dataset.

\section{\icc's relationship with interpretability}
We discuss a comparison of a traditionally interpretable machine learning model (random forest) with \icc, and we propose an architecture extension to make \icc even more interpretable (at the expense of additional complexity).

\subsection{\icc Tabular Comparison with a Traditional Interpretable Model}

In Fig. \ref{fig:icc-vs-rf}, the table (left) shows that all models, in comparison with a Random Forest model (RF), overlap in 95\% CI. The performance of a tuned RF is in line with deep learning approaches. However, the RF uses 100 trees with average depth 7.37 +/- 1.16 for Breast Cancer and 1000 trees with average depth 6.21 +/- 2.03 for Synthetic data, so is not simple to understand the decision path.

We then compare which features RF selects as important (light blue) vs. which features ICC FG selects as important (dark blue): ICC’s feature importances are much sparser.

\subsection{\icc Description of Faithfulness}
\label{app:additional-faithfulness}

As inspired from the OpenXAI benchmark \citep{agarwal2022openxai}:
\begin{itemize}[leftmargin=1em]
    \item \textbf{Feature Agreement (FA):} Computes the fraction of top-K features shared between an explanation and [the underlying data pattern].
    \item \textbf{Rank Agreement (RA):} Measures the fraction of top-K features that are not only shared but also appear in the same rank order in both the explanation and the [underlying data pattern].
\end{itemize}

We define \textbf{Ground Truth Alignment} as the fraction of features shared between an explanation and the underlying data pattern (K=total number of features). We define \textbf{Ground Truth Faithfulness} as the fraction of features shared between an explanation and the ground truth explanation (exactly what the model is using as a decision process).

As an extension of the results in Table \ref{tab:synth}, we conduct a Pairwise Rank Agreement analysis. \textbf{Pairwise Rank Agreement (PRA)} assesses whether the relative ordering of feature pairs is consistent between the explanation and underlying data pattern, computing the fraction of pairs with the same relative importance. Comparing the eight models on the synthetic data for PRA, the results are: 69.71 +/- 3.89 (ICC FG), 74.55 +/- 1.21 (ICC GR), 71.99 +/- 3.25 (SENN Features), 45.30 +/- 7.49 (SENN Concepts), 70.16 +/- 5.63 (NAM), 82.13 +/- 7.21 (IG), 73.31 +/- 1.91 (LIME), and 72.93 +/- 3.71 (SHAP). As this metric is simply a less-strict version of Rank Agreement and the order of important features is very important in downstream human-centric tasks deriving from explanations, it is not included in the main results. Notably, all 95\% CI overlap again, except for ICC GR and SENN Concepts, where SENN Concepts is significantly worse than ICC GR.

\subsection{\icc Extension for an Interpretable Discriminator Network}
\label{sec:interpretable-discriminatory}

\icc can enable an interpretable discriminator network simply by using a set of decision trees (or SVMs, LRs, any other traditionally interpretable model) instead of a neural network (FG: one model per feature or GR: one model per concept). This additional interpretability would come at the expense of model/explanation complexity and perhaps reduced accuracy as it doesn’t take into account cross-feature interactions, but would allow us to make statements like: “Concept X was chosen because feature A > value, value > B < value, C > value, D = 0, and E > value. Concept X contributes 35\% to the prediction.”

\textbf{Concrete Implementation Details}: For \icc FG, each tree would predict a binary decision between “keep this feature” or “don’t keep this feature” with the input of the full feature space. For \icc GR, each SVM would assign a score between 0 and 1 regarding whether to “keep this concept” with the input of the full feature space (the scores would then be normalized across concepts/features). These would be trained together with the predictive network and a sparsity criterion in exactly that same way as we have shown with \icc: all that changes is the choice of which model to use for the discriminator network. A simpler NN solution is to add an attention layer to the discriminator for some opaque interpretability.

However, we do not include this approach in the model as it makes both the explanation for the downstream user more complex and the model itself more complex; we choose to focus instead on the design criteria of optimizing human actionability through sparsity and simplicity.

\subsection{\icc Comparison to Extractive Rationale Methods}
\label{sec:rationale}

\textcolor{black}{In comparison to InterpretCC, extractive-rationale methods from the NLP community like FRESH (Jain et al. 2020) provide explanations that are more tailored for the text domain. However, these methods have several weaknesses, illustrated in a comparison of FRESH to InterpretCC: 1) These methods are often harder to generalize to new tasks, e.g. FRESH needs to train 3 new models sequentially instead of 1 model with two parts in parallel for every new setting. 2) There is often bias included in determining the initial importance or saliency scores (LIME, Attention, Gradients used in FRESH) instead of letting the model learn directly. 3) These models are longer, larger, and more complex to train over InterpretCC’s simpler architectures. 4) Lastly, and most importantly, methods like FRESH require the selection of a contiguous section of text as “rationale”, in contrast to InterpretCC’s mapping of words to concepts from the Dewey Decimal System (or any other grouping methodology). While this contiguous selection makes sense in the text domain, for tabular, or even time-series (beyond anomaly detection) data, this kind of explanation is not suitable or human-friendly. Therefore, we did not implement FRESH on the other modality experiments as it would be fundamentally unsuited to the data format.}

\textcolor{black}{For quantitative experiments (showcased in Table \ref{tab:all_results}, we tuned the FRESH model over rationale length and learning rate, and ran experiments with 5 random seeds with the exact architecture used in the paper. Notably, FRESH experiments use BERT models, while ICC experiments use the smaller DistillBERT model. For SST, FRESH uses a 30\% rationale size with an average accuracy of 82.05\% +/- 0.56\%, where ICC FG has 88.21\% +/- 3.41\%, ICC TopK has 92.98\% +/- 0.88\%, and ICC GR has 91.75\% +/- 1.86\%, which are significantly more performant. As SST has 7 words on average, ICC FG selects a larger explanation size, on average 3 or 4 words (indicated in Figure 2), but maintains a higher accuracy. For AG News, FRESH uses a 20\% average rational length, with performance of 88.73\% +/- 0.69\%. This is comparable to the ICC GR numbers 90.35\% +/- 1.07\% and the the ICC Top K routing approach with 87.25\% +/- 2.48\%, and more performant than the ICC FG approach 85.72\% +/- 5.31\% (although the high variation in the FG approach is to be noted), and the 95\% CIs overlap showing that the approaches are similar.}

\textcolor{black}{Other methods from the extractive-rationale community, like the methods proposed by \citep{bastings2019interpretable}, \citep{yu2019rethinking}, and \icc take fundamentally different approaches to model interpretability, each with distinct trade-offs. Bastings et al. focus on token-level rationale selection using binary gates optimized through REINFORCE, which allows for precise, fine-grained explanations but suffers from optimization challenges and limited applicability beyond text-based tasks. Yu et al. extend this idea by introducing a generator-predictor framework that enforces complementarity between selected and unselected features. While this ensures higher rationale quality, it also significantly increases model complexity and computational cost. \icc diverges by leveraging human-defined feature groups and a unified, end-to-end architecture, balancing interpretability and scalability. Unlike token-based approaches, \icc focuses on concept-level explanations that generalize across domains like tabular and time-series data, enabling more actionable insights. By emphasizing cross-feature interactions and the integration of domain knowledge, \icc differs from approaches in the extractive-rationale community.}

\begin{table}[h!]
\centering
\resizebox{\textwidth}{!}{%
\begin{tabular}{@{}lllll@{}}
\toprule
\multicolumn{1}{c}{\textbf{Aspect}} &
  \multicolumn{1}{c}{\textbf{\cite{bastings2019interpretable}}} &
  \multicolumn{1}{c}{\textbf{\cite{yu2019rethinking}}} &
  \multicolumn{1}{c}{\cite{jain2020learning}} &
  \multicolumn{1}{c}{\textbf{InterpretCC}} \\ \midrule
\textbf{Feature Selection} &
  Token-level binary gates &
  Token-level generator-predictor &
  Contiguous text spans as rationales &
  \begin{tabular}[c]{@{}l@{}}Feature-level binary gates or \\ group-level selection (domain-informed)\end{tabular} \\ \midrule
\textbf{Architecture} &
  REINFORCE optimization &
  \begin{tabular}[c]{@{}l@{}}Generator + predictor\\ (cooperative)\end{tabular} &
  \begin{tabular}[c]{@{}l@{}}Generator + predictor \\ (extractive rationales)\end{tabular} &
  \begin{tabular}[c]{@{}l@{}}Single end-to-end model \\ with routing\end{tabular} \\ \midrule
\textbf{Domain Focus} &
  Text-focused &
  Text-focused &
  Text-focused &
  General (text, tabular, time-series) \\ \midrule
\textbf{Explanation Type} &
  Token rationales &
  \begin{tabular}[c]{@{}l@{}}Token rationales \\ with complementarity\end{tabular} &
  Contiguous extractive rationales &
  \begin{tabular}[c]{@{}l@{}}Concept-based, \\ human-centric rationales\end{tabular} \\ \midrule
\textbf{\begin{tabular}[c]{@{}l@{}}Optimization \\ Complexity\end{tabular}} &
  High due to discrete variables &
  \begin{tabular}[c]{@{}l@{}}Moderate with \\ generator constraints\end{tabular} &
  \begin{tabular}[c]{@{}l@{}}Moderate due to \\ sequential training\end{tabular} &
  Low with Gumbel-Softmax routing \\ \midrule
\textbf{Main Strength} &
  Fine-grained token selection &
  Explicitly enforces rationale quality &
  Faithfulness to predictions &
  Human-friendly explanations \\ \midrule
\textbf{Main Weakness} &
  \begin{tabular}[c]{@{}l@{}}Optimization variance;\\ text-only\end{tabular} &
  \begin{tabular}[c]{@{}l@{}}High model complexity;\\ text-only\end{tabular} &
  \begin{tabular}[c]{@{}l@{}}Task-specific rationales; \\ text-only\end{tabular} &
  \begin{tabular}[c]{@{}l@{}}Requires interpretable feature \\ group design (human or LLM)\end{tabular} \\ \bottomrule
\end{tabular}}
\caption{Design comparison of \icc to Extractive Rationale Methods.}
\label{tab:comparison_methods}
\end{table}

\subsection{\icc Extension for Cross-Feature or Raw Modality Interactions}
\label{sec:graph-nn}

\textcolor{black}{Graph-based models could be very useful in the \icc architecture for raw time series data, with a natural extension in the discriminator stage. The message passing graph network and concept activation vector approach showcased with RIPPLE \citep{asadi2023ripple} could be used to define a sparse adjacency matrix as opposed to a sparse vector on the input. This could then relate to interaction-based subnetworks, referring to multiple features in each concept used in the explanation. Alternatively, graph models like RAINDROP \citep{zhanggraph} or SGP \citep{cini2023scalable} could be used in the predictive module stage, where each subnetwork is a graph model only focused on the specific modalities or features passed into the subnetwork. Using the sparseness enabled by the discriminator layers, only a few of the graph models would be activated for each point’s prediction.}

\textcolor{black}{Without this extension, it is possible \icc could be used directly on raw time series data to simply specifying the ``concepts`` as fixed or relative time intervals (analogous to anomaly detection), enabling users to identify which parts of the time-series were used in the prediction. This could answer questions like: ``Was my behavior in week 1 useful in predicting that I performed well on the exam?"). This significant time interval approach could be further extended to modalities like raw speech or video, although using modality-specific expert grouping methodologies would be preferred, as this would be more useful for downstream user actions.}

\end{document}